%% file: draft.tex
\documentclass[sigconf,authorversion=true]{acmart}

\usepackage{xcolor}
\usepackage{adjustbox}

\usepackage{colortbl}
\usepackage{tabularx}

\usepackage{booktabs} % For formal tables
\usepackage{graphicx}
\usepackage{subfig}
\usepackage[linesnumbered, vlined, ruled]{algorithm2e}
\usepackage{color, colortbl}

\definecolor{MyHiLiRow}{gray}{0.9}

\def\vector#1{\mbox{\boldmath $#1$}}

\setcopyright{rightsretained}
\copyrightyear{2021}
\acmYear{2021}
\setcopyright{acmcopyright}\acmConference[GECCO '21]{2021 Genetic and Evolutionary Computation Conference}{July 10--14, 2021}{Lille, France}
\acmBooktitle{2021 Genetic and Evolutionary Computation Conference (GECCO '21), July 10--14, 2021, Lille, France}
\acmPrice{15.00}
\acmDOI{10.1145/3449639.3459300}
\acmISBN{978-1-4503-8350-9/21/07}

\begin{document}

%% \title[Towards Exploratory Landscape Analysis for Large Scale Black-box Optimization]{Towards Exploratory Landscape Analysis for\\Large Scale Black-box Optimization}
%\title[Scaling up Exploratory Landscape Analysis]{Scaling up Exploratory Landscape Analysis for Large Scale Black-box Optimization: A Dimensionality Reduction Approach}
%\title[Towards ELA for Large Scale BBO]{Towards Exploratory Landscape Analysis for Large Scale Black-box Optimization: A Dimensionality Reduction Approach}
%\title[Towards ELA for Large Scale Black-box Numerical Optimization]{Towards Exploratory Landscape Analysis for Large Scale Black-box Numerical Optimization}
%\title[Towards ELA for Large-scale Black-box Numerical Optimization]{Towards Exploratory Landscape Analysis for Large-scale Black-box Numerical Optimization: A Dimensionality Reduction Approach}
\title[Towards ELA for Large-scale Optimization: A Dimensionality Reduction Framework]{Towards Exploratory Landscape Analysis for Large-scale Optimization: A Dimensionality Reduction Framework}

%\title[Towards ELA for Large-scale Black-box Optimization]{Towards Exploratory Landscape Analysis for Large-scale Black-box Optimization: A Dimensionality Reduction Framework}

%\title[Scaling up ELA for Large Scale Black-box Optimization]{Scaling up Exploratory Landscape Analysis for Large Scale Black-box Optimization: A Dimensionality Reduction Approach}

%% % for the double-blind review process
%% \author{\hspace{1em}}
%% \affiliation{%
%%   \institution{
%%     \hspace{1em}\\
%%     \hspace{1em}
%%   %  \hspace{1em}
%%   }
%% }
%% %\email{\hspace{1em}}

%% After acceptance
\author{Ryoji Tanabe}
\affiliation{
  \institution{
%Faculty of Environment and Information Sciences,
    Yokohama National University \& Riken AIP\\
    \city{Yokohama}
    \state{Kanagawa}
    \country{Japan}
  }
}
\email{rt.ryoji.tanabe@gmail.com}

\input{abstract.tex}

%
% The code below should be generated by the tool at
% http://dl.acm.org/ccs.cfm
% Please copy and paste the code instead of the example below. 
%



\maketitle

\input{introduction.tex}

\input{preliminaries.tex}

\input{cputime.tex}

\input{proposed_method.tex}

\input{setting.tex}

\input{results.tex}

\input{conclusion.tex}

\section*{Acknowledgments}

This work was supported by Leading Initiative for Excellent Young Researchers, MEXT, Japan.

\bibliographystyle{ACM-Reference-Format}
%\bibliography{reference}
\bibliography{long_reference} 

\end{document}

%% file: abstract.tex
\begin{abstract}

Although exploratory landscape analysis (ELA) has shown its effectiveness in various applications, most previous studies focused only on low- and moderate-dimensional problems.
Thus, little is known about the scalability of the ELA approach for large-scale optimization.
In this context, first, this paper analyzes the computational cost of features in the \texttt{flacco} package.
Our results reveal that two important feature classes (\texttt{ela\_level} and \texttt{ela\_meta}) cannot be applied to large-scale optimization due to their high computational cost.
To improve the scalability of the ELA approach, this paper proposes a dimensionality reduction framework that computes features in a reduced lower-dimensional space than the original solution space.
We demonstrate that the proposed framework can drastically reduce the computation time of \texttt{ela\_level} and \texttt{ela\_meta} for large dimensions.
In addition, the proposed framework can make the cell-mapping feature classes scalable for large-scale optimization.
Our results also show that features computed by the proposed framework are beneficial for predicting the high-level properties of the 24 large-scale BBOB functions.

\end{abstract}

%% file: introduction.tex
\section{Introduction}
\label{sec:introduction}

%The number of calls to $f$ during the optimization process should also be as small as possible.
%Here, $n$ is the dimensioanlity of a problem.
%Any explicit knowledge of $f$ is not given.
% with bound constraints.

\textit{General context.}
%\subsubsection*{General context}
We consider a \textit{noiseless} black-box optimization of an objective function $f: \mathbb{X} \rightarrow \mathbb{R}$, where $\mathbb{X} \subseteq \mathbb{R}^n$ is the $n$-dimensional solution space.
This problem involves finding a solution $\vector{x} \in \mathbb{X}$ with an objective value $f(\vector{x})$ as small as possible without any explicit knowledge of $f$.
Fitness landscape analysis \cite{PitzerA12,MalanE13} is generally used to understand the high-level properties of a problem.

% (e.g., the degree of multimodality and global structure).
%, which are beneficial for algorithm configuration and algorithm selection \cite{MunozSKH15,KerschkeHNT19}.
%The high-level properties obtained by fitness landscape analysis are beneficial for algorithm configuration and algorithm selection \cite{MunozSKH15,KerschkeHNT19}.

Exploratory landscape analysis (ELA) \cite{MersmannPT10,MersmannBTPWR11} provides a set of numerical low-level features based on a small sample of solutions.
Unlike traditional analysis methods (e.g., FDC \cite{JonesF95} and evolvability \cite{SmithHLO02}), most ELA feature values are not human-interpretable \cite{KerschkeT2019flacco}.
In the ELA approach, the extracted features are used to characterize a fitness landscape of a black-box optimization problem by machine learning.
%In the ELA approach, the extracted features are used to train and test machine learning models.
As shown in Table \ref{tab:dim_bbob}, the ELA approach has been successfully applied to various tasks.

%, including high-level property classification. and algorithm selection.
%
%Very recently, the ELA approach was also used for fitness landscape analysis of the neural architecture search problem with $n=23$ \cite{vanSteinWB20}.

Table \ref{tab:dim_bbob} shows the dimension $n$ of the BBOB functions \cite{hansen2012fun} in each previous study.
This paper denotes \textit{the noiseless BBOB functions} as \textit{the BBOB functions}.
As seen from Table \ref{tab:dim_bbob}, most previous studies\footnote{
A clear exception is \cite{BelkhirDSS17}.
For the sake of validation, $n$ was different in the training and testing phases.
The BBOB functions were used only for the training phase. 
The test functions with $n=100$ were used to validate a performance prediction model, rather than to investigate the scalability of the ELA approach.
The results in \cite{BelkhirDSS17} showed that the performance prediction model does not work well when $n$ in the testing phase is much larger than $n$ in the training phase (i.e., $n=100$).}
 focused only on low- and moderate-dimensional problems, typically with $n \leq 20$.
Thus, the scalability of the ELA approach for large-scale optimization has been poorly understood.
This limits the applicability of the ELA approach.
Although large-scale real-world problems can be found in a wide range of research areas (e.g., \cite{GohTMA15,MullerG18}), a rule of thumb is not available. %, e.g., which feature classes should be used for large-scale optimization.
Note that this situation is not unique to ELA.
Except for \cite{MorganG14}, most previous studies of fitness landscape analysis for black-box numerical optimization (e.g., \cite{MullerS11,MorganG12ppsn,ShirakawaN16}) focused only on relatively low-dimensional problems.

%Except for \cite{MorganG14}, most previous studies of fitness landscape analysis for black-box numerical optimization (e.g., \cite{MullerS11,MorganG12ppsn,ShirakawaN16}) focused only on relatively small dimensions.

\begin{table}[t]  
\setlength{\tabcolsep}{4.5pt} % Default value: 6pt
\renewcommand{\arraystretch}{0.75}
\centering
\caption{\small Dimension $n$ of the BBOB functions in selected previous studies for the following four tasks: high-level property classification (HP), algorithm selection (AS), performance prediction (PP), and per-instance algorithm configuration (PIAC).}
%      See Section \ref{sec:introduction} for the meaning of the abbreviation of each task
%{\scriptsize
{\footnotesize
%{\small
  \label{tab:dim_bbob}
\scalebox{1}[1]{ 
\begin{tabular}{cccl}
\toprule 
Ref. & Year & Task & Dimension $n$\\
\midrule
\cite{MersmannBTPWR11} & 2011 & HP & $n \in \{5, 10, 20\}$\\
\cite{MunozKH12} & 2012 & PP & $n \in \{2, 3, 5, 10, 20\}$\\
\cite{BischlMTP12} & 2012 & AS & $n \in \{5, 10, 20\}$\\
\cite{KershkePHSSGRBT14} & 2014 & HP & $n=2$\\
%\cite{MunozS15} & 2015 & HP & $n \in \{2, 3, 5, 10\}$\\
\cite{KerschkePWT15} & 2015 & HP & $n \in \{2, 3, 5, 10\}$\\
% MunozKH15 can be removed since it deals with only ic
\cite{MunozKH15} & 2015 & HP & $n \in \{2, 5, 10, 20\}$\\
\cite{KerschkePWT16} & 2016 & HP & $n \in \{2, 3, 5, 10\}$\\
\cite{BelkhirDSS16} & 2016 & PIAC & $n \in \{2, 3, 5, 10\}$\\
\raisebox{0.5em}{\cite{BelkhirDSS17}} & \raisebox{0.5em}{2017} & \raisebox{0.5em}{PIAC} & \shortstack[l]{train: $n \in \{2, 4, 5, 8, 10, 16, 20, 32, 40, 64\}$\\test: $n \in \{2, 4, 8, 10, 16, 20, 32, 40, 50, 66, 100\}$}\\
\cite{DerbelLVAT19} & 2019 & HP, AS & $n \in \{2, 3, 5\}$\\
\cite{KerschkeT19} & 2019 & AS & $n \in \{2, 3, 5, 10\}$\\
\cite{JankovicD20} & 2020 & PP & $n = 5$\\
\cite{RenauDDD20} & 2020 & HP & $n = 5$\\
\cite{EftimovPRKD20} & 2020 & HP & $n = 5$\\
\bottomrule 
\end{tabular}
}
}
\end{table}

\textit{Contributions.}
%\subsubsection*{Contributions}
In this context, first, we investigate the computation time of features in the \textsf{R}-package \texttt{flacco} \cite{KerschkeT2019flacco}, which currently provides 17 feature classes.
We use the BBOB function set and its large-scale version \cite{VarelasABHNTA20}, which consists of the 24 functions with $n \in \{20, 40, 80, 160, 320, 640\}$.
The computation time of the ELA features has been paid little attention in the literature.
While some previous studies (e.g., \cite{BischlMTP12,KerschkePWT16,DerbelLVAT19}) used the terms ``computationally cheap/expensive'' to represent a necessary sample size for ELA, this paper uses the terms only to represent the wall-clock time for computing features.
In \cite{KerschkeT2019flacco}, the computation time of the 17 feature classes in \texttt{flacco} was investigated on a problem with $n=2$.
In contrast, we analyze the influence of $n$ on the computation time of some feature classes.
The \texttt{ela\_level} and \texttt{ela\_meta} feature classes are one of the original six ELA feature classes \cite{MersmannBTPWR11}.
Some previous studies (e.g., \cite{MersmannBTPWR11,KerschkePWT15,KerschkePWT16,KerschkeT19,JankovicD20}) also reported their importance for various tasks.
However, we demonstrate that the \texttt{ela\_level} and \texttt{ela\_meta} features are not available for $n \geq 320$ due to their high computational cost.
Apart from the computational cost, as pointed out in \cite{KerschkeT2019flacco}, the four cell mapping feature classes (\texttt{cm\_angle}, \texttt{cm\_conv}, \texttt{cm\_grad}, and \texttt{gcm}) \cite{KershkePHSSGRBT14} can be applied only to small-scale problems due to their properties.
Thus, not all the 17 feature classes in \texttt{flacco} are available for large-scale optimization.

%% Although \texttt{ela\_level} and \texttt{ela\_meta} have been recognized as important feature classes since the beginning of ELA \cite{MersmannBTPWR11}, they are not extendable to large-scale optimization.

%We also report that the improved Latin hypercube sampling method \cite{BeachkofskiG02} in \texttt{flacco} is computationally expensive for a large $n$.

To improve the scalability of the ELA approach, we propose a dimensionality reduction framework that computes features in a reduced $m$-dimensional space instead of the original $n$-dimensional solution space, where $m < n$ (e.g., $m=2$ and $n=640$).
%We emphasize that dimensionality reduction is performed in \textit{the solution space, not the feature space}.
The proposed framework is inspired by dimensionality reduction strategies in Bayesian optimization \cite{WangHZMF16,HuangZSLC15,RaponiWBBD20,UllahNWMSB20}.
In this paper, we use the weighting strategy-based principal component analysis (PCA) procedure in PCA-assisted Bayesian optimization (PCA-BO) \cite{RaponiWBBD20} for dimensionality reduction.
%In this paper, we use a weighted principal component analysis (PCA) method \cite{RaponiWBBD20}, which was originally proposed for reducing the dimensionality for learning a surrogate model for computationally expensive optimization.
%
Since the proposed framework computes features in a reduced lower-dimensional space $\mathbb{R}^m$ than the original solution space $\mathbb{X} \subseteq \mathbb{R}^n$, it can drastically reduce the computation time.
In addition to the computational cost reduction, the proposed framework can make the cell mapping feature classes scalable for large-scale optimization.
We evaluate the effectiveness of the proposed framework for predicting the high-level properties of the 24 BBOB functions with up to $640$ dimensions.

%\subsubsection*{Related work}
%
%As far as we know, 
\textit{Related work.}
The proposed framework is the first attempt to compute features in a reduced $m$-dimensional space in the field of black-box numerical optimization.
We emphasize that dimensionality reduction is performed in \textit{the solution space, not the feature space}.
While some previous studies (e.g., \cite{MunozS15,ShirakawaN16,SkvorcEK20,EftimovPRKD20}) applied PCA \cite{Shlens14} to the feature space for the sake of visualization, we apply PCA to the solution space.
In the field of combinatorial optimization, some previous studies (e.g., \cite{VeerapenO18,Tayarani-NP16}) applied dimensionality reduction methods to the solution space for the sake of visualization.
In contrast, we are not interested in such a visualization.

Belkhir et al. proposed a surrogate-assisted framework that computes features based on a small-sized sample $\mathcal{X}$ \cite{BelkhirDSS16lion}.
In their framework, an augmented solution $\vector{x}^\mathrm{aug} \notin \mathcal{X}$ is evaluated by a surrogate model $M: \mathbb{R}^n \rightarrow \mathbb{R}$ instead of the actual objective function $f$.
Then, $\vector{x}^\mathrm{aug}$ is added to an augmented sample $\mathcal{X}^{\mathrm{aug}}$.
Finally, features are computed based on the union of $\mathcal{X}$ and $\mathcal{X}^{\mathrm{aug}}$.
Their framework and our framework are similar in that the features are not computed based only on the original sample $\mathcal{X}$.
However, their framework does not aim to reduce the dimension of the solution space.

We are interested in a speed-up technique at the algorithm level, not the implementation level.
One may think that the scalability issue can be addressed by reimplementing \texttt{flacco} in any compiled language (e.g., C).
However, it is not realistic to reimplement \texttt{flacco} without useful \textsf{R} libraries (e.g., mlr).
%Notice that mathematical expression of the ELA features are unknown.
Reimplementing \texttt{flacco} is also not a fundamental solution for the scalability issue.

%% The rest of this paper is organized as follows.

%\subsubsection*{Outline}
\textit{Outline.}
Section \ref{sec:preliminaries} provides some preliminaries.
Section \ref{sec:cputime} investigates the computation time of ELA features.
Section \ref{sec:proposed_method} explains the proposed framework.
Section \ref{sec:setting} explains our experimental setting.
Section \ref{sec:results} shows our analysis results.
Section \ref{sec:conclusion} concludes this paper.

%Or Data availability?
%\noindent \textbf{Reproducibility:}
%% \noindent \textbf{:}

%\subsubsection*{Code availability}
\textit{Code availability.}
The source code of the proposed framework is available at \url{https://github.com/ryojitanabe/ela_drframework}.

%\subsubsection*{Supplementary file}
\textit{Supplementary file.}
Throughout this paper, we refer to a figure and a table in the supplementary file (\url{https://ryojitanabe.github.io/pdf/gecco21-supp.pdf}) as Figure S.$*$ and Table S.$*$, respectively.

%If this paper is accepted, we will make the source code of the proposed framework publicly available on GitHub.

%If this paper is accepted for publication, we will make the source code of the proposed framework publicly available on GitHub.
%If this paper is accepted for publication, we will release the source code of the proposed framework on GitHub.

%% file: preliminaries.tex
\section{Preliminaries}
\label{sec:preliminaries}

%Here, we give some preliminaries of this work.
Section \ref{sec:testproblems} explains the BBOB functions and their high-level properties.
%Section \ref{sec:testproblems} explains the high-lelvel classification and algorithm selection.
Section \ref{sec:ela} describes ELA.
Section \ref{sec:pca_reduction} explains PCA-BO \cite{RaponiWBBD20}.

\subsection{BBOB function set}
%\subsection{High-level properties of the BBOB functions} 
\label{sec:testproblems}

%% We used the large scale version of the BBOB function set (\texttt{bbob-largescale}) 

The BBOB function set \cite{hansen2012fun} consists of the 24 functions to evaluate the performance of optimizers in terms of difficulties in real-world black-box optimization. % (e.g., separability, global structure, and variable scaling).
The functions $f_1, ...,  f_{24}$  are grouped into the following five categories: separable functions ($f_1, ...,  f_5$),  functions with low or moderate conditioning ($f_6, ..., f_9$), functions with high conditioning and unimodal ($f_{10}, ..., f_{14}$), multimodal functions with adequate global structure ($f_{15}, ..., f_{19}$), and multimodal functions with weak global structure ($f_{20}, ...,  f_{24}$).
Each BBOB function further consists many instances.
The large-scale version of the BBOB function set \cite{VarelasABHNTA20} provides problems with up to $n=640$.

Table \ref{tab:prop_bbob} shows the following seven high-level properties of the 24 BBOB functions \cite{MersmannBTPWR11}: multimodality, global structure, separability, variable scaling, search space homogeneity, basin sizes, and global to local optima contrast.
For each BBOB function, the degree of a high-level property was categorized by experts, e.g., ``none'', ``low'', ``medium'', and ``high'' for multimodality.
Since some high-level properties presented in \cite{MersmannBTPWR11} were inappropriately classified, we slightly revised them in this work.
As in \cite{KerschkePWT15}, we classified the global structure property of the 13 unimodal functions as ``strong''.
Since the Lunacek bi-Rastrigin function $f_{24}$ \cite{LunacekWS08} clearly has a multi-funnel structure, we classified its global structure property as ``deceptive''.
As in \cite{MersmannPTBW15}, we revised the separability property of $f_3$, $f_6$, and $f_7$.
According to the definitions in \cite{hansen2012fun}, we revised the variable scaling and global to local optima contrast properties of some functions.

%% Since only $f_1$ is labeld as ``None'' in the variable scaling property, it is imposible to estimate the true property of $f_1$ for the LOPO-CV.

\definecolor{c1}{RGB}{192,192,192}
  
\begin{table}[t]
  \newcommand{\rotdeg}{70}
\setlength{\tabcolsep}{4.25pt} % Default value: 6pt
\renewcommand{\arraystretch}{0.85}  
  \centering
  \caption{\small High-level properties of the 24 BBOB functions.
    The properties revised from \cite{MersmannPT10} are highlighted.}
%The properties revised from \cite{MersmannPT10} are highlighted.}  
\label{tab:prop_bbob}
%% {\scriptsize
%% {\footnotesize
{\small
\scalebox{1}[1]{ 
\begin{tabular}{cccccccc}
  \toprule
%  &  \rotatebox{\rotdeg}{multimodality} &  \rotatebox{\rotdeg}{global structure} &  \rotatebox{\rotdeg}{separability} &  \rotatebox{\rotdeg}{variablescaling} &  \rotatebox{\rotdeg}{homogeneity} &  \rotatebox{\rotdeg}{basinsizes} &  \rotatebox{\rotdeg}{glcontrast}\\
% &  \rotatebox{\rotdeg}{multim.} &  \rotatebox{\rotdeg}{gl-struc.} &  \rotatebox{\rotdeg}{separ.} &  \rotatebox{\rotdeg}{scaling} &  \rotatebox{\rotdeg}{homog.} &  \rotatebox{\rotdeg}{basin} &  \rotatebox{\rotdeg}{gl-cont.}\\  
 &  multim. &  gl-struc. &  separ. &  scaling &  homog. &  basin &  gl-cont.\\  
  \toprule
$f_{1}$ &  none &  \cellcolor{c1}strong &  high &  none &  high &  none &  none\\
$f_{2}$ &  none &  \cellcolor{c1}strong &  high &  high &  high &  none &  none\\
$f_{3}$ &  high &  strong &  \cellcolor{c1}high &  low &  high &  low &  low\\
$f_{4}$ &  high &  string &  high &  low &  high &  med. &  low\\
$f_{5}$ &  none &  \cellcolor{c1}strong &  high &  \cellcolor{c1}low &  high &  none &  none\\
\midrule
$f_{6}$ &  none &  \cellcolor{c1}strong &  \cellcolor{c1}none &  low &  med. &  none &  none\\
$f_{7}$ &  none &  \cellcolor{c1}strong &  \cellcolor{c1}none &  low &  high &  none &  none\\
$f_{8}$ &  low &  \cellcolor{c1}strong &  none &  \cellcolor{c1}low &  med. &  low &  low\\
$f_{9}$ &  low &  \cellcolor{c1}strong &  none &  \cellcolor{c1}low &  med. &  low &  low\\
\midrule
$f_{10}$ &  none &  \cellcolor{c1}strong &  none &  high &  high &  none &  none\\
$f_{11}$ &  none &  \cellcolor{c1}strong &  none &  high &  high &  none &  none\\
$f_{12}$ &  none &  \cellcolor{c1}strong &  none &  high &  high &  none &  none\\
$f_{13}$ &  none &  \cellcolor{c1}strong &  none &  med. &  med. &  none &  none\\
$f_{14}$ &  none &  \cellcolor{c1}strong &  none &  med. &  med. &  none &  none\\
\midrule
$f_{15}$ &  high &  strong &  none &  low &  high &  low &  low\\
$f_{16}$ &  high &  med. &  none &  med. &  high &  med. &  low\\
$f_{17}$ &  high &  med. &  none &  low &  med. &  med. &  high\\
$f_{18}$ &  high &  med. &  none &  high &  med. &  med. &  high\\
$f_{19}$ &  high &  strong &  none &  \cellcolor{c1}low &  high &  low &  low\\
\midrule
$f_{20}$ &  med. &  dec. &  none &  \cellcolor{c1}low &  high &  low &  \cellcolor{c1}high\\
$f_{21}$ &  med. &  none &  none &  med. &  high &  med. &  low\\
$f_{22}$ &  low &  none &  none &  \cellcolor{c1}high &  high &  med. &  \cellcolor{c1}high\\
$f_{23}$ &  high &  none &  none & \cellcolor{c1}low &  high &  low &  low\\
$f_{24}$ &  high &   \cellcolor{c1}dec. &  none &  low &  high &  low &  low\\
  \toprule
\end{tabular}
}
}
\end{table}

\subsection{ELA}
\label{sec:ela}

%% (\textbf{TODO: re-organize the structure of this section})
%Below, we explain how to apply ELA to a problem.
%ELA \cite{MersmannBTPWR11} produces a set of numerical features to capture the fitness landscape of a problem or to select the best optimizer from an algorithm portfolio automatically.
%Below, we explain how the ELA approach produces numerical features on a problem instance.

ELA \cite{MersmannBTPWR11} produces a set of numerical features based on a set of $l$ solutions $\mathcal{X} = \{\vector{x}_i\}^l_{i=1}$, where $\mathcal{X}$ is called the initial sample.
%Here, $\vector{x}$ and $\mathcal{X}$ are denoted as an observation and the initial sample, respectively.
In general, $\mathcal{X}$ is randomly generated in the $n$-dimensional solution space $\mathbb{X} \subseteq \mathbb{R}^n$ by a sampling method (e.g., Latin hypercube sampling).
%In \cite{KerschkePWT16}, $s = 50$ is sufficient.
%In most previous studies, $l$ was set to $l = s \times n$ so that $l$ linearly increases with respect to $n$.
%In \cite{KerschkePWT16}, $s = 50$ is sufficient.
%The previous work \cite{KerschkePWT16} showed that $s = 50$ is sufficient.
Then, the objective value $f(\vector{x})$ is calculated for each solution $\vector{x} \in \mathcal{X}$.
Let $\mathcal{Y}$ be $\{f(\vector{x}_i)\}^l_{i=1}$.
We also denote a pair of $\mathcal{X}$ and $\mathcal{Y}$ as a data set $\mathcal{D}$.
Finally, an ELA feature maps $\mathcal{D}$ to a numerical value.

% $\mathbb{R}$.
%An ELA feature $F$ produces a numerical feature value based on $\mathcal{D}$, i.e., $F: \mathcal{D} \rightarrow \mathbb{R}$.

%maps $\mathcal{D}$ to a numerical feature value $\mathbb{R}$.
%Finally,
%An ELA feature maps $\mathcal{D}$ to a numerical feature value $\mathbb{R}$.
%Finally, a single feature maps $\mathcal{D}$ to the corresponding numerical feature value $\mathbb{R}$.
%% Next, an ELA feature $\{\vector{X}, f(\vector{X})\} \mapsto \mathbb{R}$ is calculated.

%% the initial sample $\vector{X} = \{\vector{x}_1, ..., \vector{x}_q\}$ with the size $q$ is generated.
%% In general, the initial sample $\vector{X}$ is generated using Latin hypercube sampling (LHS).
%% In \texttt{flacco}, the improved Latin hypercube sampling method (ILHS) \cite{BeachkofskiG02} is used as a default sampling method.
%% At the beginning of the procedure, ILHS randomly generate a small number of solutions $\vector{X}$.
%% Then, new solutions are sequentially added to $\vector{X}$ in a greedy manner so that the nearest-neighbor distance between solutions in the resulting $\vector{X}$ is close to an ideal distance $q/\sqrt[n]{q}$ (\textbf{TODO: re-check the procedure of ILHS}).
%% Then, $\vector{X}$ is evaluated by the objective function $f$.

%% Although determining a proper choice of $q$ is not easy \cite{KerschkePWT16,JankovicD20}, $q$ is generally set to $q = s\: n$ so that $q$ is scale to $n$.
%% Thus, the sampling size $q$ linearly increasses with respect to $n$.
%% Here, the previous work \cite{KerschkePWT16} shows that $s = 50$ is enough.

The \textsf{R}-package \texttt{flacco} \cite{KerschkeT2019flacco} is generally used to compute features.
%Although ELA started from the six feature classes \cite{MersmannBTPWR11}, a new feature class has continuously been proposed in the literature.
%Currently, \texttt{flacco} provides 17 feature classes, which consist of 343 features.
%% Currently, \texttt{flacco} provides 343 features, which are grouped into 17 feature classes.
%% Thus, each feature class consists of more than one feature.
%
Currently, \texttt{flacco} provides 17 feature classes.
Each feature class consists of more than one feature.
For example, the \texttt{ela\_distr} class provides the five features: \texttt{skewness}, \texttt{kurtosis}, \texttt{number\_of\_peaks}, \texttt{costs\_fun\_evals}, and \texttt{costs\_runtime}.
%, which map a pair of $\mathcal{X}$ and $\mathcal{Y}$ to a numerical feature value $\mathbb{R}$, respectively.

%\texttt{gcm} \cite{KershkePHSSGRBT14} & GCM & 75 & & \checkmark & \\

%For exmample, the \texttt{ela\_distr} class provides the following five features: \texttt{ela\_distr.skewness}, \texttt{ela\_distr.kurtosis}, \texttt{ela\_distr.number\_of\_peaks}, \texttt{ela\_distr.costs\_fun\_evals}, and \texttt{ela\_distr.costs\_runtime}.

%Table \ref{tab:flacco_features} shows 14 ELA feature classes provided by \texttt{flacco} \cite{KerschkeT2019flacco}, where \texttt{flacco} is an \textsf{R}-package to compute the ELA feature values.
%The original six ELA feature classes \cite{MersmannBTPWR11} are convexity (\texttt{ela\_conv}), curvature (\texttt{ela\_curv}), local search (\texttt{ela\_local}), \texttt{ela\_distr}, levelset (\texttt{ela\_level}, and \texttt{ela\_meta}.

Table \ref{tab:flacco_features} shows 14 feature classes provided by \texttt{flacco}.
%Note that the number of features does not indicate the importance of the corresponding class. 
%Below, we brifly explain the 17 ELA feature classes provided by \texttt{flacco}.
The original six ELA  feature classes \cite{MersmannBTPWR11} are \texttt{ela\_conv}, \texttt{ela\_curv}, \texttt{ela\_local}, \texttt{ela\_distr}, \texttt{ela\_level}, and \texttt{ela\_meta}.
%They are currently called classical ELA features.
As in most recent studies (see Table \ref{tab:dim_bbob}), we do not consider \texttt{ela\_conv}, \texttt{ela\_curv}, and \texttt{ela\_local}.
This is because they require additional function evaluations independently from the initial sample $\mathcal{X}$.

%% Since \texttt{ela\_conv}, \texttt{ela\_curv}, and \texttt{ela\_local} need additional function evaluations independently from the initial sample $\mathcal{X}$, most recent studies did not use them.
%% For this reason, we do not consider \texttt{ela\_conv}, \texttt{ela\_curv}, and \texttt{ela\_local}.

The \texttt{ela\_distr} features are with regard to the distribution of $\mathcal{Y}$.
The \texttt{ela\_level} class splits $\mathcal{D}$ into binary classes based on a pre-defined threshold value. % either of the 10\%, 20\%, and 50\%-quantile of $\mathcal{Y}$.
%Then, three classifiers (LDA, QDA, and MDA) are used to predict which class an objective value $f(\vector{x}) \in \mathcal{Y}$ should be categorized.
Then, three classifiers (LDA, QDA, and MDA) are used to classify an objective value $f(\vector{x}) \in \mathcal{Y}$.
Each \texttt{ela\_level} feature represents the mean misclassification errors of a classifier over a 10-fold cross-validation.
The \texttt{ela\_meta} class fits linear and quadratic regression models to $\mathcal{D}$.
The \texttt{ela\_meta} features are the model-fitting results, e.g., the adjusted $R^2$.

The \texttt{nbc} features aim to detect funnel structures by using the nearest-better clustering method \cite{Preuss12}, which was originally proposed for multimodal optimization.
The \texttt{disp} class is an extension of the dispersion metric \cite{LunacekW06}, which is one of the most representative metrics for landscape analysis.
The \texttt{ic} class provides features regarding the information content of fitness sequences \cite{MunozKH15}.

%The three feature classes (\texttt{basic}, \texttt{limo}, and \texttt{pca}) are based on simple ideas.
As the name suggests, the \texttt{basic} features provide basic information about $\mathcal{D}$, e.g., the size of $\mathcal{X}$ and the maximum value of $\mathcal{Y}$.
The \texttt{limo} class fits a linear model to $\mathcal{X}$ as the explanatory variables and $\mathcal{Y}$ as the dependent variables.
The \texttt{pca} class applies PCA to $\mathcal{X}$ and $\mathcal{D}$ separately.
The \texttt{pca} features are based on the covariance and the correlation matrix calculated by PCA \cite{Shlens14}.

The cell mapping feature classes (\texttt{cm\_angle}, \texttt{cm\_grad}, \texttt{cm\_conv}, \texttt{gcm}, and \texttt{bt}) \cite{KershkePHSSGRBT14,KerschkeT2019flacco} discretize the $n$-dimensional solution space into $b^n$ cells, where $b$ is the number of blocks for each dimension.
The minimum value of $b$ is $3$ \cite{KerschkeT2019flacco}.
Note that the number of cells $b^n$ grows exponentially with respect to $n$.
For example, $b^{n} \approx 3.5 \times 10^9$ for $b=3$ and $n=20$.
For this reason, the cell mapping feature classes can be applied only to low-dimensional problems \cite{KerschkeT2019flacco}.
%
%In the default setting, $b$ is set to 3.
The \texttt{cm\_angle} and \texttt{cm\_grad} features are mainly based on the angle and gradient of the position of solutions in a cell, respectively.
The \texttt{cm\_conv} class aims to characterize the convexity of the fitness landscape.
Most \texttt{gcm} features are based on the transition probability to move from a cell to one of its neighbors.
The \texttt{bt} features aim to capture the local optimal solutions and the ridges of the fitness landscape based on the leaves of the tree and the branching nodes.

\begin{table}[t]
\centering
\caption{\small 14 feature classes provided by \texttt{flacco}, except for \texttt{ela\_conv}, \texttt{ela\_curv}, and \texttt{ela\_local}.}
  % (except for \texttt{ela\_conv}, \texttt{ela\_curv}, and \texttt{ela\_local}).}
\label{tab:flacco_features}
\renewcommand{\arraystretch}{0.85}  
%% {\scriptsize
%% {\footnotesize
{\small
\scalebox{1}[1]{ 
\begin{tabular}{llc}
  \toprule
Feature class & Name & Num. features\\  
\toprule
%% \texttt{ela\_conv} \cite{MersmannBTPWR11} & ELA convexity & 6 & \checkmark &  & \\
%% %\midrule
%% \texttt{ela\_curv} \cite{MersmannBTPWR11} & ELA curvature & 26 & \checkmark &  & \\
%% %\midrule
%% \texttt{ela\_local} \cite{MersmannBTPWR11} & ELA level search & 16 & \checkmark &  & \\
%% \midrule
%% #expensive features: 'ela_conv', 'ela_curv', 'ela_local'
%% # only for 2 or 3 dim: 'bt', 
%% # cell mapping features: 'cm_angle', 'cm_conv', 'cm_grad', 'gcm',
\texttt{ela\_distr} \cite{MersmannBTPWR11} & $y$-distribution & 5\\
\texttt{ela\_level} \cite{MersmannBTPWR11} & levelset & 20\\
\texttt{ela\_meta} \cite{MersmannBTPWR11} & meta-model & 11\\
%\midrule
\texttt{nbc} \cite{KerschkePWT15} & nearest better clustering (NBC) & 7\\
\texttt{disp} \cite{KerschkePWT15} & dispersion & 18\\
\texttt{ic} \cite{MunozKH15} & information content & 7\\
%\midrule
\texttt{basic} \cite{KerschkeT2019flacco} & basic & 15\\
\texttt{limo} \cite{KerschkeT2019flacco} & linear model & 14\\
\texttt{pca} \cite{KerschkeT2019flacco} & principal component analysis & 10\\
\midrule
\texttt{cm\_angle} \cite{KershkePHSSGRBT14} & cell mapping angle & 10\\
\texttt{cm\_conv} \cite{KershkePHSSGRBT14} & cell mapping convexity & 6\\
\texttt{cm\_grad} \cite{KershkePHSSGRBT14} & cell mapping gradient homog. & 6\\
%\midrule
\texttt{gcm} \cite{KershkePHSSGRBT14} & generalized cell mapping & 75\\
%\midrule
\texttt{bt} \cite{KerschkeT2019flacco} & barrier tree & 90\\
\toprule
\end{tabular}
}
}
\end{table}

\subsection{PCA-BO}
\label{sec:pca_reduction}

Bayesian optimization is an efficient sequential model-based approach for computationally expensive optimization \cite{ShahriariSWAF16}.
The Gaussian process regression (GPR) model is generally used in Bayesian optimization.
However, it is difficult to apply the GPR model to large-scale problems.
This is because the GPR model requires high computational cost as the dimension increases.
One promising way to address this issue is the use of dimensionality reduction \cite{WangHZMF16,HuangZSLC15,RaponiWBBD20,UllahNWMSB20}.
A similar approach has been adopted in the filed of evolutionary computation (e.g., \cite{LiuZG14}).
%
%% Roughly speaking, a dimensionality reduction method is applied to a set of $n$-dimensional solutions $\mathcal{X}$. % which consists of $n$-dimensional solutions.
%% As a result, each solution $\vector{x} \in \mathcal{X}$ is mapped to an $m$-dimensional vector $\hat{\vector{x}}$, where $m \ll n$.
%% Then, the GPR model is fitted to $\hat{\mathcal{X}}$, instead of the original $\mathcal{X}$.

%Then, the GPR model is fitted to $\hat{\mathcal{X}} = \{\hat{\vector{x}}\}^l_{i=1}$, instead of the original $\mathcal{X}$.

%PCA is a linear transformation method taht maps the $n$-dimensioanl solution space $\mathbb{X} \subseteq \mathbb{R}^n$ to the $m$-dimensional space $\mathbb{R}^m$.
%proposed in the field of Bayesian optimization.%

%% PCA-assisted Bayesian optimization (PCA-BO) \cite{RaponiWBBD20} uses PCA for dimensionality reduction.
%% PCA is a linear transformation technique that reduces the dimensionality of a high-dimensional dataset while minimizing information loss \cite{Shlens14}.
%% In a nutshell, PCA applies an orthogonal transformation to a dataset to obtain a new coordinate system.
%% Then, the principal components are sorted based on their variances in decreasing order.
%% Finally, PCA selects the first $m$ principal components that keep the most important information about the dataset.
%% PCA-BO also uses a weighting strategy to incorporate the information about the objective values into solutions so that a better solution is treated as more important than other solutions.

PCA-assisted Bayesian optimization (PCA-BO) \cite{RaponiWBBD20} uses PCA to reduce the original dimension $n$ to a lower dimension $m$ ($m<n$).
PCA-BO also uses a weighting strategy to incorporate the information about the objective values into solutions so that a better solution is treated as more important than other solutions.

%In addion to PCA, PCA-BO uses a weighting strategy to incorporate the information about the objective values $\mathcal{Y}$ into a set of solutions $\mathcal{X}$ so that a better solution is treated as more important than other solutions.

%PCA-BO uses PCA for dimensionality reduction of solutions.
%
%Let $\mathcal{Y}$ be the objective values of $\mathcal{X}$.

%$\hat{\mathcal{X}} = \{\hat{\vector{x}}\}^l_{i=1}$

At the beginning of each iteration, PCA-BO re-scales a set of $l$ solutions found so far $\mathcal{X} = \{\vector{x}\}^l_{i=1}$ by the weighting strategy.
First, all the solutions in $\mathcal{X}$ are ranked based on their objective values.
For $i \in \{1, ..., l\}$, a weight value $w_i$ is assigned to the $i$-th solution $\vector{x}_i \in \mathcal{X}$ based on its rank $r_i$ as follows: $w_i =  \tilde{w}_i / \sum^l_{j=1} \tilde{w}_j$, where $\tilde{w}_i =  \ln{l} - \ln{r_i}$.
Then, $\vector{x}_i$ is re-scaled as follows: $\bar{\vector{x}}_i = w_i (\vector{x}_i - \vector{m})$, where $\vector{m} = 1/l\sum_{\vector{x} \in \mathcal{X}} \vector{x}$ is the mean vector of $\mathcal{X}$.
After the re-scaled version $\bar{\mathcal{X}}$ of $\mathcal{X}$ is obtained by the weighting strategy, PCA-BO applies PCA to $\bar{\mathcal{X}}$.
The dimensionality of each point in $\bar{\mathcal{X}}$ is reduced from $n$ to $m$.
In other words, PCA maps an $n$-dimensional point $\bar{\vector{x}} \in \mathbb{R}^n$ to an $m$-dimensional point $\hat{\vector{x}} \in \mathbb{R}^m$.
Then, PCA-BO fits a GPR model to $\hat{\mathcal{X}} = \{\hat{\vector{x}}\}^l_{i=1}$ and searches a candidate solution that maximizes the acquisition function in $\mathbb{R}^m$.

%% file: cputime.tex
%\section{Feature computation time in ELA}
\section{Computational cost issue in ELA}
\label{sec:cputime}

Here, we investigate the computation time of the nine non-cell mapping ELA feature classes in Table \ref{tab:flacco_features} on problems with up to $n=640$.
Since the cell mapping features can be computed only for small dimensions, we do not consider them in this section.

%Apart from the calculation time of the features, we point out that the improved Latin hypercube sampling method (IHS) \cite{BeachkofskiG02} in \texttt{flacco} is time-consuming for large-scale optimization.

%We do not show detailed results of IHS, but our preliminary results show that IHS requires approximately 6.6 hours for sampling $50 \times n$ solutions on a problem instance for $n=160$.

Apart from the features, we point out that the improved Latin hypercube sampling method (IHS) \cite{BeachkofskiG02} in \texttt{flacco} is time-consuming for large dimensions.
Our results show that IHS requires approximately 6.6 hours for sampling $50 \times n$ solutions on a problem instance for $n=160$.
The procedure of IHS did not finish within one day for $n \geq 320$.
For details, see Figure S.1.
This is because IHS calculates the Euclidean distance between points every time IHS adds a new point.
%Thus, IHS is not available for large-scale optimization. % due to its high computational cost.
For this reason, we use Latin hypercube sampling (LHS) instead of IHS throughout this paper.

% due to space limitation

\subsection{Experimental setup}
\label{sec:cputime_setting}

We conducted all experiments on a workstation with an Intel(R) 40-Core Xeon Gold 6230 (20-Core$\times 2$) 2.1GHz and 384GB RAM using Ubuntu 18.04.
We used Python version 3.8 and \textsf{R} version  3.6.3.
We used the Python interface of \texttt{flacco} (\texttt{pflacco}), which is available at \url{https://github.com/Reiyan/pflacco}.
We believe that the time to call a \texttt{flacco} function from \texttt{pflacco} is negligible when $n$ is large enough.
We used the BBOB function set \cite{hansen2012fun} for $n \in \{2, 3, 5, 10\}$ and its large-scale version \cite{VarelasABHNTA20} for $n \in \{20, 40, 80, 160, 320, 640\}$.
Both function sets are available in the COCO platform  \cite{HansenARMTB21}.
We performed 31 independent runs on the first instance of $f_1$ to measure the average time for computing features.
We used \texttt{lhs} in the pyDOE package to generate the initial sample $\mathcal{X}$.
As recommended in \cite{KerschkePWT16}, we set $|\mathcal{X}|$ to $50 \times n$ (i.e., $l = 50 \times n$ in Section \ref{sec:pca_reduction}).
We did not take into account the time to generate $\mathcal{X}$ and evaluate their objective values $\mathcal{Y}$.
Thus, we measured only the time to compute features. % based on $\mathcal{X}$.

%\subsection{Results}
\subsection{Feature computation time in ELA}
\label{sec:cputime_ela}

\begin{figure}[t]
   \centering
   \includegraphics[width=0.47\textwidth]{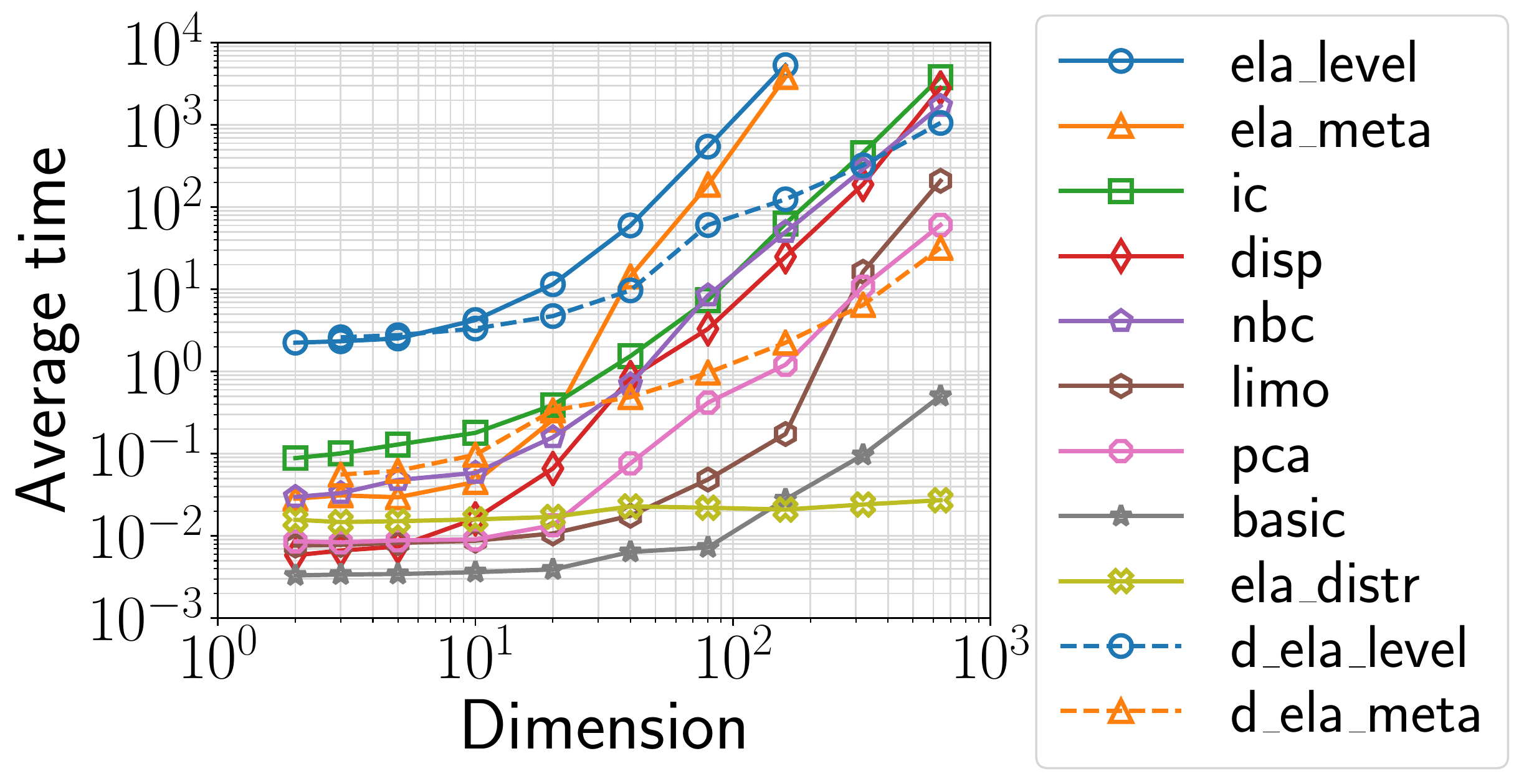}
   \caption{
     Average feature computation time (sec) on the first instance of $f_1$ with $n \in \{2, 3, 5, 10, 20, 40, 80, 160, 320, 640\}$.
   }
   \label{fig:cputime_features}
\end{figure}

Figure \ref{fig:cputime_features} shows the average computation time of each feature class over 31 runs.
We explain results of \texttt{d\_ela\_level} and \texttt{d\_ela\_meta} in Section \ref{sec:runtime_redu}.
It seems that the computation time of all the feature classes (except for \texttt{ela\_distr}) increases exponentially with respect to $n$.
Note that the computation time is influenced by both $n$ and $|\mathcal{X}|$.
Figure S.2 shows results when fixing $|\mathcal{X}|$ to $100$.
Figure S.3 shows results when fixing $n$ to $2$.
Although setting $|\mathcal{X}|$ to a small constant number (i.e., $|\mathcal{X}|=100$) can speed up the computation of features, the resulting features are likely to be ineffective.

%% Figures S.2 and S.3 show how $n$ and $|\mathcal{X}|$ infuluence the computation time, respectively.
%% The In Figures S.2

%As seen from Figure \ref{fig:cputime_features}, the computation time of \texttt{basic} is lowest for $n \leq 80$.
As seen from Figure \ref{fig:cputime_features}, the computation time of \texttt{basic} is lowest for $n \leq 80$.
For $n \geq 160$, \texttt{ela\_distr} is the fastest feature class in terms of the computation time, where the number of the \texttt{ela\_distr} features is only five (see Table \ref{tab:flacco_features}).
Since the \texttt{ela\_distr} features are based only on the objective values $\mathcal{Y}$, its computational cost depends only on $|\mathcal{Y}|$.
Since \texttt{limo} and \texttt{pca} are based on relatively simple linear models, their computational cost is acceptable even for $n = 640$.
The single-run computation of the \texttt{limo} and \texttt{pca} features took approximately 3.5 minutes and 1.1 minutes for $n=640$, respectively.
For $n \geq 40$, \texttt{ic}, \texttt{nbc}, and \texttt{disp} perform relatively similar in terms of the computation time.
This is because the three feature classes calculate the distance between solutions in $\mathcal{X}$, where the distance calculation time depends on $n$ and $|\mathcal{X}|$.

The computational cost of the \texttt{ela\_meta} features is relatively low for $n \leq 20$, but it increases drastically for $n \geq 40$.
This is mainly because fitting a quadratic model in \texttt{ela\_meta} is time-consuming for large dimensions.
Clearly, the computation of \texttt{ela\_level} is the most expensive for any $n$.
This is because the \texttt{ela\_level} features are based on results of the three classifiers in a 10-fold cross-validation as explained in Section \ref{sec:ela}.
As shown in Figure \ref{fig:cputime_features}, the single-run computation of the \texttt{ela\_meta} and \texttt{ela\_level} features took approximately 1.1 hours and 1.5 hours for $n=160$, respectively.
We could not measure the computation time of the \texttt{ela\_level} and \texttt{ela\_meta} features for $n \geq 320$ because their single-run computation for $n = 320$ did not finish within 3 days.

%% file: proposed_method.tex
%\section{Proposed dimensionality reduction approach}
\section{Proposed framework}
\label{sec:proposed_method}

Our results in Section \ref{sec:cputime} showed that the two important feature classes (\texttt{ela\_level} and \texttt{ela\_meta}) are not available for large-scale optimization due to their time-consuming process.
Apart from that, as pointed out in \cite{KerschkeT2019flacco}, the cell mapping feature classes can be applied only to low-dimensional problems.
To improve the scalability of these feature classes, we propose a simple dimensionality reduction framework.
Let $\mathcal{X} = \{\vector{x}_i\}^l_{i=1}$ be a set of $l$ $n$-dimensional solutions, i.e., the initial sample.
Let also $\mathcal{Y} = \{f(\vector{x}_i)\}^l_{i=1}$ be a set of $l$ objective values.
As explained in Section \ref{sec:ela}, the features are computed based on $\mathcal{X}$ and $\mathcal{Y}$.
However, as discussed above, some features cannot be computed in a practical time when $n$ is large.

%However, as discussed above, some features are not computable when $n$ is large.

%computable dimension, feasible dimension, admissible dimension
%of the simplest

One simple way of addressing this issue is to reduce the original dimension $n$ to a lower dimension $m$ as in dimensionality reduction strategies in Bayesian optimization.
Previous studies (see Section \ref{sec:pca_reduction}) show that a dimensionality reduction strategy can effectively reduce the computational cost for fitting a surrogate model for large-scale optimization.
Based on these promising results, we suppose that some (not all) features computed in the reduced $m$-dimensional space can substitute for their original versions.

%Based on these promising results, we suppose that a fitness landscape of the reduced $m$-dimensional space is not totally different from that of the original $n$-dimensional solution space.

%Otherwise, the surrogate model fitted in the $m$-dimensional space performs a poor prediction.
%For these reasons, it is reasnable to substiute features extracted in the $m$-dimensional space for fearues extracted in the original $n$-dimensional space.

In the proposed framework, first, a dimensionality reduction method is applied to $\mathcal{X}$. % which consists of $n$-dimensional solutions.
As a result, each solution $\vector{x} \in \mathbb{R}^n$ in $\mathcal{X}$ is mapped to a point $\hat{\vector{x}} \in \mathbb{R}^m$, where $m < n$.
Let $\hat{\mathcal{X}} = \{\hat{\vector{x}}_i\}^l_{i=1}$ be a set of $l$ $m$-dimensional points transformed from $\mathcal{X}$.
For $i \in \{1, ..., l\}$, the objective value $f(\vector{x}_i)$ is assigned to $\hat{\vector{x}_i}$ without any change, i.e., $\hat{\mathcal{Y}} := \mathcal{Y}$.
Thus, the proposed framework does not require additional function evaluations.
Finally, features are computed based on $\hat{\mathcal{X}}$ and $\hat{\mathcal{Y}}$ instead of $\mathcal{X}$ and $\mathcal{Y}$.

Although any dimensionality reduction method can be integrated into the proposed framework, this paper uses the weighting strategy-based PCA approach in PCA-BO \cite{RaponiWBBD20} (see Section \ref{sec:pca_reduction}).
PCA is a simple linear transformation technique, but it is one of the most representative dimensionality reduction methods \cite{vanderMaatenPH08}.
For this reason, PCA is a reasonable first choice.
Note that the PCA procedure in PCA-BO is independent from the \texttt{pca} feature class.

%explained in Section \ref{sec:ela}.

One advantage of the proposed framework is that it can speed up the computation of time-consuming features since the features are computed in a reduced $m$-dimensional space. % than the original $n$-dimensional solution space.
Another advantage is that the proposed framework can make the cell mapping feature classes scalable for large-scale optimization by setting $m$ to a small value (e.g., $m=2$).
One disadvantage is that features computed by the proposed framework may be misleading.
This is because it is impossible for the proposed framework to exactly extract all information about the original $n$-dimensional problem.
For this reason, it is expected that the original features are more effective than their dimensionality reduction versions when they are available.

%% file: setting.tex
\section{Experimental setup}
\label{sec:setting}

This section explains the experimental setup.
Unless explicitly noted, the computational environment is the same as in Section \ref{sec:cputime_setting}.
As in most previous studies (e.g., \cite{MersmannBTPWR11,KershkePHSSGRBT14,KerschkePWT16,DerbelLVAT19}), we used an off-the-shelf random forest classifier \cite{Breiman01} for predicting the seven high-level properties of the 24 BBOB functions (see Table \ref{tab:prop_bbob}).
We aim to examine the effectiveness of a feature set for the high-level property classification task rather than to achieve high accuracy by using a state-of-the-art classifier (e.g., deep neural network).
We employed the scikit-learn implementation of random forest \cite{PedregosaVGMTGBPWDVPCBPD11}.
We set the number of trees to $1\,000$.
According to the recommendation in \cite{KerschkePWT16}, we set the size of the initial sample $|\mathcal{X}|$ ($=l$) to $50 \times n$.

Below, Section \ref{sec:feature_sets} explains feature sets investigated in Section \ref{sec:results}.
Section \ref{sec:cross-validation} describes the cross-validation procedure in this work.

\subsection{Feature sets}
\label{sec:feature_sets}

%As in Section We used the \texttt{flacco} and \texttt{pflacco}

Table \ref{tab:feature_set} shows five feature sets investigated in this work.
For a feature class computed by the proposed dimensionality reduction framework, we add a prefix ``\texttt{d\_}'' to its original name.
%For example, \texttt{d\_ela\_level} means the dimensionality reduction version of \texttt{ela\_level}.
For dimensionality reduction, we employed the implementation of PCA-BO provided by the authors of \cite{RaponiWBBD20} (\url{https://github.com/wangronin/Bayesian-Optimization}).
We set the reduced dimension $m$ to $2$.
%, where $m=2$ is the minimumu value.

Our results in Section \ref{sec:cputime_ela} show that the following seven feature classes are relatively computationally cheap for large dimensions: \texttt{ela\_distr}, \texttt{basic}, \texttt{ic}, \texttt{disp}, \texttt{nbc}, \texttt{pca}, and \texttt{limo}.
We denote a set of the \underline{seven} computationally \underline{c}heap feature classes as C7.
We consider C7 as a base-line feature set for analysis.
Morgan and Gallagher demonstrated that the performance of the dispersion metric \cite{LunacekW06} for large dimensions ($n \leq 200$) can be improved by normalization \cite{MorganG14}.
One may think that the \texttt{disp} features with normalization is more effective for large dimensions.
However, our preliminary results showed that normalization does not strongly affect the \texttt{disp} features.
This is mainly because the \texttt{disp} features are already normalized by the mean or the median of the dispersion value of the whole sample $\mathcal{X}$.
Of course, further investigation is needed.
%, but it is out of the scope of this paper.

%Section \ref{sec:comparison_dimredu_e2} applies the proposed dimensionalitiy reductino appraoch to the two computaitonally expensive feature classes (\texttt{ela\_meta} and \texttt{ela\_level}).
C7-E2 is a C7 with the \underline{two} computationally \underline{e}xpensive feature classes (\texttt{ela\_level} and \texttt{ela\_meta}), which can be computed only for $n \leq 160$ due to their time-consuming process.
C7-D2 is a C7 with the \underline{d}imensionality reduction versions of the \underline{two} expensive feature classes.
We analyze the effectiveness of the proposed framework by comparing C7-D2 with C7 and C7-E2.

%Section \ref{sec:comparison_dimredu_e2} applies the proposed framework to the four cell mapping feature classes (\texttt{gcm}, \texttt{cm\_angle}, \texttt{cm\_conv}, and \texttt{cm\_grad}), where they are denoted as \texttt{d\_gcm}, \texttt{d\_cm\_angle}, \texttt{d\_cm\_conv}, and \texttt{d\_cm\_grad}, respectively.
C7-C4 is a C7 with the \underline{four} \underline{c}ell mapping feature classes, which are available only for $n \leq 5$.
Since \texttt{bt} can be computed only for $n=2$, we did not use \texttt{bt}.
C7-D4 is a C7 with the \underline{d}imensionality reduction versions of the \underline{four} cell mapping feature classes.
Before computing the cell mapping features, we normalized each point $\hat{\vector{x}} $ in $ \hat{\mathcal{X}}$ into the range $[0,1]^m$ using the minimum and maximum values in $\hat{\mathcal{X}}$ for each dimension.

%No previoius study has also repoted the effectiveness of \texttt{bt}.
%For these reasons, 

For each $n$, we deleted a feature that takes the same value on all instances (e.g., \texttt{disp.costs\_fun\_evals} and \texttt{basic.dim}).
As in \cite{BelkhirDSS17,DerbelLVAT19,RenauDDD20}, we did not perform feature selection because it deteriorated the performance of a classifier in our preliminary experiment.
A similar observation was reported in \cite{JankovicED21}.
This may be due to a \textit{leave-one-problem-out cross-validation} (LOPO-CV) \cite{KershkePHSSGRBT14,DerbelLVAT19} (see the next section), where the prediction model is validated on an unseen function.
Although an analysis of cross-validation methods is beyond the scope of this paper, an in-depth investigation is needed.

\begin{table}[t]
  \renewcommand{\arraystretch}{0.7}
  \centering
  \caption{\small Five feature sets investigated in this work.}
%{\scriptsize
%{\footnotesize
{\small
  \label{tab:feature_set}
\scalebox{1}[1]{ 
\begin{tabular}{ll}
\toprule
Name & Feature classes\\
\midrule
%
%C4 & $\{$ \texttt{basic}, \texttt{ela\_distr}, \texttt{pca}, \texttt{limo} $\}$\\
%C7 & $\{$\texttt{basic}, \texttt{ela\_distr}, \texttt{pca}, \texttt{limo}, \texttt{ic}, \texttt{disp}, \texttt{nbc}$\}$\\
C7 & $\{$\texttt{ela\_distr}, \texttt{basic}, \texttt{ic}, \texttt{disp}, \texttt{nbc}, \texttt{pca}, \texttt{limo}$\}$\\
\midrule
C7-E2 & C7 $\cup$ $\{$\texttt{ela\_level}, \texttt{ela\_meta}$\}$\\
C7-D2 & C7 $\cup$ $\{$\texttt{d\_ela\_level}, \texttt{d\_ela\_meta}$\}$\\
\midrule
C7-C4 & C7 $\cup$ $\{$\texttt{gcm}, \texttt{cm\_angle}, \texttt{cm\_conv}, \texttt{cm\_grad}$\}$\\
C7-D4 & C7 $\cup$ $\{$\texttt{d\_gcm}, \texttt{d\_cm\_angle}, \texttt{d\_cm\_conv}, \texttt{d\_cm\_grad}$\}$\\
%% \midrule
\bottomrule
\end{tabular}
}
}
\end{table}

\subsection{Cross-validation procedure}
\label{sec:cross-validation}

The ultimate goal of high-level property classification is to predict a high-level property of an unseen real-world problem.
Thus, we need to evaluate the unbiased performance of a classifier on unseen test problem instances.
For this reason, we adopted LOPO-CV \cite{KershkePHSSGRBT14,DerbelLVAT19}.
In LOPO-CV, a 24-fold cross-validation is performed on the 24 BBOB functions with each dimension $n$.
Let $\mathcal{I}_{i}$ be a set of problem instances of a function $f_i$, where $i \in \{1, ..., 24\}$.
Since each BBOB function consists of 15 instances in the COCO platform, $|\mathcal{I}_{i}| = 15$ for any $f_i$.
While we used only the first instance of $f_1$ to measure the wall-clock time in Section \ref{sec:cputime}, we consider all the 15 instances of all the 24 BBOB functions in Sections \ref{sec:comparison_dimredu_e2} and \ref{sec:comparison_redu_features}.
In the $i$-th fold, $\mathcal{I}_{i}$ (15 instances) is used in the testing phase.
Thus, $\mathcal{I}_{1} \cup \cdots \cup \mathcal{I}_{24} \setminus \mathcal{I}_{i}$ ($23 \times 15 = 345$ instances) are used in the training phase.
Since the 24 BBOB functions have totally different properties from each other (see Table \ref{tab:prop_bbob}), problem instances used in the training and testing phases are also different in LOPO-CV.
For this reason, high-level property classification in LOPO-CV is very challenging.

We also evaluated the performance of a classifier by using a \textit{leave-one-instance-out cross-validation} (LOIO-CV) \cite{DerbelLVAT19}.
In LOIO-CV, a 15-fold cross-validation is performed on the 15 problem instances.
In the $i$-th fold, the 24 $i$-th problem instances are used in the testing phase.
Thus, the remaining $14 \times 24 = 336$ instances are used in the training phase.
Since problem instances used in the training and testing phases are very similar,  LOIO-CV is much easier than LOPO-CV.
In fact, our preliminary results showed that some classifiers could perfectly predict some high-level properties of the 24 BBOB functions in LOIO-CV (i.e., accuracy $ = 1$).
For this reason, we consider only the more challenging LOPO-CV in this paper.

%% file: results.tex
\section{Results}
\label{sec:results}

%% This section investigates the effectiveness of the proposed dimensionality reduction framework.
%% For the sake of simplicity, we refer to ``a random forest classification model using a feature set $\mathcal{F}$'' as ``$\mathcal{F}$''.
%% For example, we denote ``the accuracy of a random forest classifier using C7-D2'' as ``the accuracy of C7-D2''.

First, Section \ref{sec:runtime_redu} demonstrates that the proposed framework can reduce the computational cost of the two expensive feature classes (\texttt{ela\_meta} and \texttt{ela\_level}) for large dimensions.
Then, Section \ref{sec:comparison_dimredu_e2} examines the effectiveness of the five feature sets in Table \ref{tab:feature_set} for predicting the high-level properties of the 24 large-scale BBOB functions.
%Section \ref{sec:m_redu} examines the influence of the reduced dimension $m$ in PCA on the performance of a classifier.
Finally, Section \ref{sec:comparison_redu_features} analyzes the similarity between the original features and their dimensionality reduction versions.

For the sake of simplicity, we refer to ``a random forest classification model using a feature set $\mathcal{F}$'' as ``$\mathcal{F}$''.
For example, we denote ``the accuracy of a random forest classifier using C7-D2'' as ``the accuracy of C7-D2''.

%Computation time of the proposed framework
\subsection{Computation time reduction}
\label{sec:runtime_redu}

%\texttt{d\_ela\_level}

Figure \ref{fig:cputime_features} (see Section \ref{sec:cputime_ela}) shows the average time for computing the \texttt{d\_ela\_level} and \texttt{d\_ela\_meta} features.
%Figure \ref{fig:cputime_features} (see Section \ref{sec:cputime_ela}) shows the computation time of the \texttt{d\_ela\_level} and \texttt{d\_ela\_meta} features.
In Figure \ref{fig:cputime_features}, the computation time includes the PCA-BO procedure for dimensionality reduction.
Since the proposed framework cannot be applied to problems with $n \leq m$, we do not show results of the \texttt{d\_ela\_level} and \texttt{d\_ela\_meta} features for $n=2$.
Figure S.4 shows results of the dimensionality reduction versions of the four cell mapping feature classes in C7-D4.
They perform relatively similar to \texttt{pca} and \texttt{d\_ela\_meta} for $n=640$.

%The experimental setting is the same as in Section \ref{sec:cputime_setting}.

%Since we set $m$ to 2, 
%in terms of the computaitonal cost

%As seen from Figure \ref{}, we can say that the runtime of the dimensionlaity reduction verisons of the the four cell mapping feature classes is acceptable even for $n=640$.

As shown in Figure \ref{fig:cputime_features}, the computation time of \texttt{d\_ela\_level} is slightly higher than that of \texttt{ela\_level} for $n \in \{3, 5\}$.
This is because the dimensionality reduction procedure is more time-consuming than the feature computation procedure for small dimensions.
In contrast, \texttt{d\_ela\_level} is much faster than \texttt{ela\_level} in terms of the computation time as the dimension increases.
For example, while the computation of \texttt{ela\_level} requires approximately 1.5 hours for $n = 160$, that of \texttt{d\_ela\_level} requires only approximately 2.1 minutes.
Notably, the computation of \texttt{d\_ela\_level} is cheaper than that of the three distance-based feature classes (\texttt{ic}, \texttt{disp}, and \texttt{nbc}) for $n=640$.
Results of \texttt{d\_ela\_meta} is similar to the results of \texttt{d\_ela\_level} discussed above.
The computation of \texttt{d\_ela\_meta} is much cheaper than that of \texttt{ela\_meta} for $n \geq 40$.
Interestingly, \texttt{d\_ela\_meta} is faster than \texttt{pca} for $n \geq 320$ in terms of the computation time.
%We believe that this is due to the difference between the execution time of Python and \textsf{R}, where we used the Python implementation of PCA-BO for dimensionality reduction.
As seen from Figure \ref{fig:cputime_features}, the computation of \texttt{d\_ela\_meta} is faster than that of \texttt{limo} for $n \geq 320$.
This result indicates that fitting a quadratic model in a reduced $m$-dimensional space can be faster than fitting a linear model in the original $n$-dimensional space as $n$ increases.
In summary, our results show that the proposed framework can effectively reduce the computation time of \texttt{ela\_meta} and \texttt{ela\_level} for large dimensions.

% dimensioanlity reduction%

%% The calculation of \texttt{d\_ela\_meta} requires more computational cost than that of \texttt{ela\_meta} for $n \in \{3, 5, 10, 20\}$.
%% In contrast, \texttt{d\_ela\_meta} is much faster than \texttt{ela\_meta} for $n \geq 40$ in terms of the runtime.
%% Interestingly, the calculation time of \texttt{d\_ela\_meta} is faster than that of \texttt{limo} and \texttt{pca}.

% \definecolor{c1}{RGB}{192,192,192}
%% \definecolor{c2}{RGB}{220,220,220}

\definecolor{c1}{RGB}{150,150,150}
\definecolor{c2}{RGB}{220,220,220}

%% \definecolor{c1}{RGB}{192,192,192}
%% \definecolor{c2}{RGB}{0,0,220}

\begin{table*}[t]
  \renewcommand{\arraystretch}{0.8} 
\centering
  \caption{\small Average accuracy of C7, C7-E2, and C7-D2 on the 24 BBOB functions with $n \in \{2, 3, 5, 10, 20, 40, 80, 160, 320, 640\}$.}
  \label{tab:hpc_accuracy_lopocv_e2}  
%{\scriptsize
{\footnotesize
%{\small
%%%%%%%%%%%
\subfloat[Multimodality]{
\begin{tabular}{cccc}
\toprule
& C7 & C7-E2 & C7-D2\\  
\midrule
$2$ & 0.642 & \cellcolor{c1}0.703 & Na\\
$3$ & 0.569 & \cellcolor{c1}0.597 & \cellcolor{c2}0.578\\
$5$ & 0.536 & \cellcolor{c1}0.625 & \cellcolor{c2}0.622\\
$10$ & 0.522 & \cellcolor{c2}0.631 & \cellcolor{c1}0.647\\
$20$ & 0.531 & \cellcolor{c1}0.725 & \cellcolor{c2}0.639\\
$40$ & 0.556 & \cellcolor{c1}0.689 & \cellcolor{c2}0.617\\
$80$ & 0.600 & \cellcolor{c1}0.783 & \cellcolor{c2}0.622\\
$160$ & 0.522 & \cellcolor{c1}0.650 & \cellcolor{c2}0.561\\
$320$ & 0.544 & Na & \cellcolor{c1}0.578\\
$640$ & 0.514 & Na & \cellcolor{c1}0.583\\
\toprule
\end{tabular}
}
%%%%%%%%%%%
\subfloat[Global structure]{
\begin{tabular}{cccc}
\toprule
& C7 & C7-E2 & C7-D2\\  
\midrule
$2$ & 0.758 & \cellcolor{c1}0.786 & Na\\
$3$ & \cellcolor{c2}0.792 & \cellcolor{c1}0.797 & 0.778\\
$5$ & \cellcolor{c2}0.769 & \cellcolor{c1}0.772 & 0.761\\
$10$ & \cellcolor{c2}0.708 & \cellcolor{c1}0.750 & 0.692\\
$20$ & \cellcolor{c2}0.714 & \cellcolor{c1}0.742 & 0.697\\
$40$ & 0.711 & \cellcolor{c2}0.711 & \cellcolor{c1}0.711\\
$80$ & 0.697 & \cellcolor{c2}0.697 & \cellcolor{c1}0.697\\
$160$ & \cellcolor{c2}0.647 & \cellcolor{c1}0.650 & 0.631\\
$320$ & 0.664 & Na & \cellcolor{c1}0.667\\
$640$ & 0.667 & Na & \cellcolor{c1}0.678\\
\toprule
\end{tabular}
}
%%%%%%%%%%%
\subfloat[Separability]{
\begin{tabular}{cccc}
\toprule
& C7 & C7-E2 & C7-D2\\  
\midrule
$2$ & 0.756 & \cellcolor{c1}0.803 & Na\\
$3$ & \cellcolor{c2}0.789 & \cellcolor{c1}0.856 & 0.786\\
$5$ & 0.758 & \cellcolor{c1}0.864 & \cellcolor{c2}0.761\\
$10$ & \cellcolor{c2}0.758 & \cellcolor{c1}0.828 & 0.753\\
$20$ & 0.708 & \cellcolor{c1}0.808 & \cellcolor{c2}0.744\\
$40$ & \cellcolor{c2}0.767 & \cellcolor{c1}0.836 & 0.758\\
$80$ & \cellcolor{c2}0.703 & \cellcolor{c1}0.775 & 0.686\\
$160$ & \cellcolor{c2}0.697 & \cellcolor{c1}0.753 & 0.667\\
$320$ & \cellcolor{c1}0.722 & Na & 0.669\\
$640$ & \cellcolor{c1}0.725 & Na & 0.669\\
\toprule
\end{tabular}
}
%%%%%%%%%%%
\subfloat[Variable scaling]{
\begin{tabular}{cccc}
\toprule
& C7 & C7-E2 & C7-D2\\  
\midrule
$2$ & 0.581 & \cellcolor{c1}0.586 & Na\\
$3$ & \cellcolor{c2}0.611 & 0.583 & \cellcolor{c1}0.611\\
$5$ & \cellcolor{c1}0.628 & \cellcolor{c2}0.600 & 0.594\\
$10$ & 0.508 & \cellcolor{c2}0.556 & \cellcolor{c1}0.572\\
$20$ & 0.539 & \cellcolor{c2}0.578 & \cellcolor{c1}0.583\\
$40$ & 0.539 & \cellcolor{c1}0.581 & \cellcolor{c2}0.556\\
$80$ & 0.542 & \cellcolor{c1}0.583 & \cellcolor{c2}0.581\\
$160$ & 0.558 & \cellcolor{c2}0.583 & \cellcolor{c1}0.583\\
$320$ & 0.547 & Na & \cellcolor{c1}0.583\\
$640$ & 0.542 & Na & \cellcolor{c1}0.542\\
\toprule
\end{tabular}
}
\\
%%%%%%%%%%%
\subfloat[Homogeneity]{
\begin{tabular}{cccc}
\toprule
& C7 & C7-E2 & C7-D2\\  
\midrule
$2$ & 0.633 & \cellcolor{c1}0.636 & Na\\
$3$ & \cellcolor{c1}0.664 & 0.608 & \cellcolor{c2}0.608\\
$5$ & \cellcolor{c2}0.647 & 0.578 & \cellcolor{c1}0.647\\
$10$ & \cellcolor{c2}0.722 & 0.658 & \cellcolor{c1}0.769\\
$20$ & \cellcolor{c1}0.772 & 0.731 & \cellcolor{c2}0.731\\
$40$ & \cellcolor{c1}0.761 & 0.700 & \cellcolor{c2}0.744\\
$80$ & \cellcolor{c2}0.703 & 0.639 & \cellcolor{c1}0.742\\
$160$ & \cellcolor{c2}0.656 & 0.625 & \cellcolor{c1}0.683\\
$320$ & 0.719 & Na & \cellcolor{c1}0.761\\
$640$ & 0.772 & Na & \cellcolor{c1}0.781\\
\toprule
\end{tabular}
}
%%%%%%%%%%%
\subfloat[Basin size]{
\begin{tabular}{cccc}
\toprule
& C7 & C7-E2 & C7-D2\\  
\midrule
$2$ & 0.450 & \cellcolor{c1}0.492 & Na\\
$3$ & \cellcolor{c2}0.544 & \cellcolor{c1}0.547 & 0.522\\
$5$ & 0.508 & \cellcolor{c1}0.628 & \cellcolor{c2}0.522\\
$10$ & 0.494 & \cellcolor{c2}0.608 & \cellcolor{c1}0.636\\
$20$ & 0.467 & \cellcolor{c1}0.647 & \cellcolor{c2}0.558\\
$40$ & 0.531 & \cellcolor{c1}0.617 & \cellcolor{c2}0.614\\
$80$ & 0.464 & \cellcolor{c1}0.650 & \cellcolor{c2}0.597\\
$160$ & 0.472 & \cellcolor{c2}0.578 & \cellcolor{c1}0.614\\
$320$ & 0.408 & Na & \cellcolor{c1}0.508\\
$640$ & 0.456 & Na & \cellcolor{c1}0.528\\
\toprule
\end{tabular}
}
%%%%%%%%%%%
\subfloat[Global to local optima cont.]{
\begin{tabular}{cccc}
\toprule
& C7 & C7-E2 & C7-D2\\  
\midrule
$2$ & 0.572 & \cellcolor{c1}0.642 & Na\\
$3$ & \cellcolor{c2}0.586 & \cellcolor{c1}0.600 & 0.572\\
$5$ & 0.497 & \cellcolor{c1}0.669 & \cellcolor{c2}0.536\\
$10$ & 0.542 & \cellcolor{c1}0.656 & \cellcolor{c2}0.619\\
$20$ & 0.544 & \cellcolor{c1}0.728 & \cellcolor{c2}0.617\\
$40$ & 0.533 & \cellcolor{c1}0.708 & \cellcolor{c2}0.597\\
$80$ & 0.558 & \cellcolor{c1}0.703 & \cellcolor{c2}0.619\\
$160$ & 0.553 & \cellcolor{c1}0.636 & \cellcolor{c2}0.622\\
$320$ & 0.550 & Na & \cellcolor{c1}0.647\\
$640$ & 0.572 & Na & \cellcolor{c1}0.614\\
\toprule
\end{tabular}
}
%%%%%%%%%%%
\subfloat[Overall average]{
\begin{tabular}{cccc}
\toprule
& C7 & C7-E2 & C7-D2\\  
\midrule
$2$ & 0.627 & \cellcolor{c1}0.664 & Na\\
$3$ & \cellcolor{c2}0.651 & \cellcolor{c1}0.656 & 0.637\\
$5$ & 0.621 & \cellcolor{c1}0.677 & \cellcolor{c2}0.635\\
$10$ & 0.608 & \cellcolor{c2}0.669 & \cellcolor{c1}0.670\\
$20$ & 0.611 & \cellcolor{c1}0.708 & \cellcolor{c2}0.653\\
$40$ & 0.628 & \cellcolor{c1}0.692 & \cellcolor{c2}0.657\\
$80$ & 0.610 & \cellcolor{c1}0.690 & \cellcolor{c2}0.649\\
$160$ & 0.587 & \cellcolor{c1}0.639 & \cellcolor{c2}0.623\\
$320$ & 0.594 & Na & \cellcolor{c1}0.631\\
$640$ & 0.607 & Na & \cellcolor{c1}0.628\\
\toprule
\end{tabular}
}
%%%%%%%%%%%
}
\end{table*}

\begin{table}[t]
  \renewcommand{\arraystretch}{0.85} 
  \centering
\caption{\small Overall average accuracy of C7, C7-C4, and C7-D4 for predicting the seven high-level properties of the BBOB functions.}
  \label{tab:hpc_accuracy_lopocv_c4}
  %{\scriptsize
{\footnotesize
%{\small
  %%%%%%%%%%%
\begin{tabular}{cccc}
\toprule
& C7 & C7-C4 & C7-D4\\  
\midrule
$2$ & 0.627 & \cellcolor{c1}0.634 & Na\\
$3$ & \cellcolor{c2}0.651 & \cellcolor{c1}0.661 & 0.648\\
$5$ & \cellcolor{c2}0.621 & \cellcolor{c1}0.632 & 0.616\\
$10$ & 0.608 & Na & \cellcolor{c1}0.628\\
$20$ & 0.611 & Na & \cellcolor{c1}0.623\\
$40$ & 0.628 & Na & \cellcolor{c1}0.632\\
$80$ & 0.610 & Na & \cellcolor{c1}0.618\\
$160$ & 0.587 & Na & \cellcolor{c1}0.593\\
$320$ & 0.594 & Na & \cellcolor{c1}0.600\\
$640$ & 0.607 & Na & \cellcolor{c1}0.614\\
%
%% The first submitted version. The results of C7-C4 are incorrect. We found that this version of C7-C4 included the ela_meat and ela_level features. For this reason, this version of C7-C4 achives a slightly better accuracy than the correct version of C7-C4. However, this mistake does not affect our conclusion.
%% $2$ & 0.627 & \cellcolor{c1}0.671 & Na\\
%% $3$ & \cellcolor{c2}0.651 & \cellcolor{c1}0.662 & 0.648\\
%% $5$ & \cellcolor{c2}0.621 & \cellcolor{c1}0.679 & 0.616\\
%% $10$ & 0.608 & Na & \cellcolor{c1}0.628\\
%% $20$ & 0.611 & Na & \cellcolor{c1}0.623\\
%% $40$ & 0.628 & Na & \cellcolor{c1}0.632\\
%% $80$ & 0.610 & Na & \cellcolor{c1}0.618\\
%% $160$ & 0.587 & Na & \cellcolor{c1}0.593\\
%% $320$ & 0.594 & Na & \cellcolor{c1}0.600\\
%% $640$ & 0.607 & Na & \cellcolor{c1}0.614\\
\toprule
\end{tabular}
%%%%%%%%%%%
}
\end{table}

\subsection{Effectiveness of features computed by the proposed framework}
\label{sec:comparison_dimredu_e2}

\subsubsection{Results of C7, C7-E2, and C7-D2}

%Table \ref{tab:hpc_accuracy_lopocv_e2} shows results of the average classification accuracy of C7, C7-E2, and C7-D2 over the different folds for each $n$.

%We made a single model for each of $9$ dimensionanlty.
%unbiased

Tables \ref{tab:hpc_accuracy_lopocv_e2}(a)--(g) show the average accuracy of C7, C7-E2, and C7-D2 for the seven high-level properties, respectively.
%Recall that ``C7'' is synonymous with ``a random forest classification model using the feature set C7'' (see the beginning of Section \ref{sec:results}).
Table \ref{tab:hpc_accuracy_lopocv_e2}(h) also shows the overall average.
In Table \ref{tab:hpc_accuracy_lopocv_e2}, ``Na'' means that the corresponding feature set is not available for a given $n$ due to its time-consuming process.
The best and second best data are highlighted by \adjustbox{margin=0.1em, bgcolor=c1}{dark gray} and \adjustbox{margin=0.1em, bgcolor=c2}{gray}, respectively.
When only two data are available, only the best data is highlighted by \adjustbox{margin=0.1em, bgcolor=c1}{dark gray}.
Note that we built one random forest model for each fold, each feature set, each classification task, and each dimension.
Tables S.1 (a)--(g) also show the standard deviation of the accuracy.
Due to LOPO-CV, the standard deviation is large.

%As mentioned in the beginning of this section, for the sake of brevity, we synonymously refer to ``a random forest classification model using a feature set $\mathcal{F}$'' as ``$\mathcal{F}$''.

As shown in Table \ref{tab:hpc_accuracy_lopocv_e2}, C7-E2 performs the best in terms of the accuracy in most cases for $n \leq 160$, except for the prediction of the homogeneity property.
These results indicate the importance of the \texttt{ela\_meta} and \texttt{ela\_level} features to predict a high-level property of an unseen problem.
However, C7-E2 is not available for $\geq 320$ due to the high-computational cost of \texttt{ela\_meta} and \texttt{ela\_level}.
In contrast, C7-D2 achieves better accuracy than C7 in most cases (especially for $n \geq 320$), except for the prediction of the separability property.
These results indicate that the \texttt{d\_ela\_meta} and \texttt{d\_ela\_level} features can be substituted for their original versions for large-scale optimization.
The \texttt{d\_ela\_meta} and \texttt{d\_ela\_level} features are particularly beneficial for predicting the multimodality and basin size properties (see Tables \ref{tab:hpc_accuracy_lopocv_e2}(a) and (f)).
Since the proposed framework reduces the dimensionality of each point in the initial sample $\mathcal{X}$ from $n$ to $m$, it is unsurprising that the \texttt{d\_ela\_meta} and \texttt{d\_ela\_level} features do not provide meaningful information about separability.
Consequently, for the prediction of the separability property, C7-D2 performs worse than C7 in terms of accuracy due to the misleading features (see Table \ref{tab:hpc_accuracy_lopocv_e2}(c)).
As already discussed in Section \ref{sec:proposed_method}, this is one disadvantage of the proposed framework.
%These results indicate the effectivenes of \texttt{d\_ela\_meta} and \texttt{d\_ela\_level} for large-scale optimization.
%, which their original \texttt{ela\_meta} and \texttt{ela\_level} cannot be applied to.

%

%Most previous studies in Table \ref{tab:dim_bbob} evaluted the performance of prediction models based on results on problems with a single dimensionality or summrized results on problmes with multiple dimensionalityes.

%To the best of our knowledge, 

The influence of the dimension $n$ on the performance of the ELA approach has not been well analyzed in the literature.
Intuitively, it is expected that a prediction model performs poorly as the dimension $n$ increases.
Roughly speaking, the results in Tables \ref{tab:hpc_accuracy_lopocv_e2}(a), (b), (c), and (d) are consistent with our intuition.
In contrast, the results in Tables \ref{tab:hpc_accuracy_lopocv_e2}(e), (f), and (g) show that the accuracy of C7, C7-E2, and C7-D2 is improved as $n$ increases.
For example, the accuracy of C7 for predicting the homogeneity property is 0.633 and 0.772 for $n=2$ and $n=640$, respectively.
This may be because the size of the initial sample $\mathcal{X}$ increases linearly with respect to $n$, where we set $|\mathcal{X}|$ to $50 \times n$.
The large-sized $\mathcal{X}$ can possibly be beneficial to capture the degree of the three properties (homogeneity, basin size, and global to local optima contrast) independently from $n$.

%A further analysis is necessary.

\subsubsection{Results of C7, C7-C4, and C7-D4}
\label{sec:res_cell}

Table \ref{tab:hpc_accuracy_lopocv_c4} shows the overall average accuracy of C7, C7-C4, and C7-D4.
Table S.2 shows detailed results.
As seen from Table \ref{tab:hpc_accuracy_lopocv_c4}, C7-C4 performs the best in terms of accuracy for $n \leq 5$.
%C7-C4 is also competitive with C7-E2 for $n \leq 5$ (see Table \ref{tab:hpc_accuracy_lopocv_c4} and Tables \ref{tab:hpc_accuracy_lopocv_e2}(h)).
Although the cell mapping features are available only for $n \leq 5$, the proposed framework can improve their scalability for $n > 5$.
As shown in the results of C7-D4, the cell mapping features computed by the proposed framework are helpful for high-level property classification for $n \geq 10$.
However, C7-D4 performs poorly for small dimensions similar to C7-D2.
Thus, it is better to use the proposed framework only for large dimensions.

%Based on this observation, it is better to use the proposed framework only for large dimensions.
%Recall that we designed the proposed framework for large-scale optimization rather than small-scale optimization.

%As demonstrated in \cite{KershkePHSSGRBT14}, the cell mapping features are helpful for high-level property classification.

%The poor performance of C7-C4 for the predciton of variable scaling for $n \in \{3, 5\}$ is consisitent with \cite{KershkePHSSGRBT14}.

\subsubsection{Feature importance}
\label{sec:feature_importance}

%Our results show the effectiveness of C7-D2 and C7-D4 for large dimensions.
We investigate which features are particularly important in C7-D2 and C7-D4.
Table \ref{tab:avg_ranks_features} shows the average rankings of the features computed by the proposed framework for $n = 640$.
Results for all dimensions are relatively similar.
%Since results for all dimensions are relatively similar, we show the results only for $n = 640$.
First, we ranked the features based on their impurity-based feature importance values computed by the random forest for each fold, where we used the \texttt{feature\_importances\_} attribute of scikit-learn.
Then, we calculated the average rankings of the features over the 24-folds and all the seven classification tasks for each dimension.
Table \ref{tab:avg_ranks_features} shows only the results of top 10 features.
Tables S.3--S.6 show the results of all features for all dimensions.

%The number of features in C7-D2 and C7-D4 were

As seen from Table \ref{tab:avg_ranks_features}(a), the seven \texttt{d\_ela\_meta} features are ranked as more important than the \texttt{d\_ela\_level} features.
For this reason, one may think that the \texttt{d\_ela\_level} features can be removed from C7-D2.
However, we observed that removing the \texttt{d\_ela\_level} features from C7-D2 degrades its performance.
As shown in Table \ref{tab:avg_ranks_features}(b), the \texttt{d\_cm\_angle} features are more important than the other cell mapping features.
In contrast to C7-D2, all the cell mapping features always rank low.
For example, the average ranking of \texttt{d\_cm\_angle.y\_ratio\_best2worst.sd} is 43.2.
However, as demonstrated in Section \ref{sec:res_cell},  the cell mapping features computed by the proposed framework are beneficial for predicting the high-level properties of the BBOB functions.
This result indicates the difficulty of feature selection in the ELA approach.

%Altough Table \ref{tab:avg_ranks_features}(b) shows that the cell-mapping features computd by the proposed framework are not so important, this does not mean 
%  \caption{\small Average ranks of the features in C7-D2 and C7-D4. In the table, \texttt{dem}, \texttt{del}, \texttt{dca}, \texttt{dcc}, \texttt{dcg}, \texttt{dg} stand for \texttt{d\_ela\_meta}, \texttt{d\_ela\_level}, \texttt{d\_cm\_angle}, \texttt{d\_cm\_conv}, \texttt{d\_cm\_grad}, and \texttt{d\_gcm}, respectively.

\begin{table}[t]
\setlength{\tabcolsep}{2pt} % Default value: 6pt
\renewcommand{\arraystretch}{0.85}  
  \centering
  \caption{\small Average rankings of the \texttt{d\_ela\_meta} (\texttt{dem}), \texttt{d\_ela\_level} (\texttt{del}), \texttt{d\_cm\_angle} (\texttt{dca}), \texttt{d\_cm\_conv} (\texttt{dcc}), \texttt{d\_cm\_grad} (\texttt{dcg}), and \texttt{d\_gcm} (\texttt{dg}) features in C7-D2 and C7-D4.
  }
  \label{tab:avg_ranks_features}
%  {\scriptsize
{\footnotesize
%{\small
  %%%%%%%%%%%
\subfloat[C7-D2]{
  \begin{tabular}{lllllllllllllllllll}
  \toprule
Feature & Rank\\    
\midrule
\texttt{dem.lin\_simple.coef.max} & 10.4\\
\texttt{dem.lin\_simple.intercept} & 14.0\\
\texttt{dem.lin\_simple.coef.min} & 22.7\\
\texttt{dem.quad\_simple.adj\_r2} & 22.8\\
\texttt{dem.quad\_w\_interact.adj\_r2} & 23.2\\
\texttt{dem.lin\_simple.adj\_r2} & 23.9\\
\texttt{dem.lin\_w\_interact.adj\_r2} & 25.5\\
\texttt{del.mmce\_qda\_25} & 29.7\\
\texttt{del.mmce\_lda\_50} & 29.7\\
\texttt{del.mmce\_mda\_50} & 30.1\\
\toprule
\end{tabular}
  }
  %%%%%%%%%%%
  %%%%%%%%%%%
\subfloat[C7-D4]{
  \begin{tabular}{lc}
    \toprule
Feature & Rank\\    
\midrule
\texttt{dca.y\_ratio\_best2worst.sd} & 43.2\\
\texttt{dca.angle.mean} & 44.7\\
\texttt{dcc.costs\_runtime} & 47.1\\
\texttt{dcg.mean} & 50.5\\
\texttt{dg.near.costs\_runtime} & 50.8\\
\texttt{dca.angle.sd} & 51.1\\
\texttt{dca.dist\_ctr2worst.mean} & 51.4\\
\texttt{dca.dist\_ctr2best.mean} & 52.2\\
\texttt{dca.y\_ratio\_best2worst.mean} & 52.6\\
\texttt{dca.dist\_ctr2best.sd} & 52.9\\
\toprule
\end{tabular}
  }
  %%%%%%%%%%%

}
\end{table}

\subsubsection{Analysis of $m$ in the proposed framework}
\label{sec:m_redu}

The reduced dimension $m$ can be viewed as a control parameter in the proposed framework.
Although we set $m$ to 2 in this section, we here investigate the influence of $m$ on the performance of the ELA approach.

Table \ref{tab:hpc_different_m} shows results of C7-D2 and C7-D4 with different $m$.
Tables S.7 and S.8 show detailed results.
Since the cell mapping features can be computed for $2 \leq n \leq 5$, we set $m$ to 2, 3, and 5 for C7-D4.
Table \ref{tab:hpc_different_m}(a) and (b) indicate that a suitable $m$ depends on $n$ and a feature set.
Roughly speaking, for C7-D2, $m=2$ is suitable for $n \leq 320$, but $m=10$ is the best setting for $n=640$.
Based on this observation, we suggest the use of $m=2$ for C7-D2, but it may be better to set $m$ to a large value (e.g., $m \in \{5, 10\}$) for $n \geq 640$.
For C7-D4, it is difficult to give a general conclusion, but any $m \in \{2, 3, 5\}$ can be a good first choice.

\begin{table}[t]
  \setlength{\tabcolsep}{4.8pt} % Default value: 6pt
\renewcommand{\arraystretch}{0.85}    
  \centering
\caption{\small Overall average accuracy of C7-D2 and C7-D4 with different $m$ for $n \in \{5, 10, 20, 40, 80, 160, 320, 640\}$.}
  \label{tab:hpc_different_m}
  %{\scriptsize
{\footnotesize
%        {\small
  %%%%%%%%%%%
\subfloat[C7-D2]{  
  \begin{tabular}{cccccc}
\toprule
%& $m=1$ & $m=2$ & $m=3$ & $m=5$ & $m=10$\\
& $1$ & $2$ & $3$ & $5$ & $10$\\
\midrule
%% $2$ & Na & Na & Na & Na\\
%% $3$ & 0.637 & Na & Na & Na\\
$5$ & \cellcolor{c1}0.640 & 0.635 & \cellcolor{c2}0.639 & Na & Na\\
$10$ & 0.654 & \cellcolor{c1}0.670 & \cellcolor{c2}0.665 & 0.643 & Na\\
$20$ & \cellcolor{c2}0.655 & 0.653 & \cellcolor{c1}0.662 & 0.652 & 0.644\\
$40$ & \cellcolor{c2}0.656 & \cellcolor{c1}0.657 & 0.649 & 0.652 & 0.648\\
$80$ & 0.639 & \cellcolor{c1}0.649 & \cellcolor{c2}0.646 & 0.635 & 0.636\\
$160$ & 0.617 & \cellcolor{c2}0.623 & 0.620 & \cellcolor{c1}0.626 & 0.615\\
$320$ & 0.621 & \cellcolor{c1}0.631 & \cellcolor{c2}0.629 & 0.628 & 0.624\\
$640$ & 0.626 & 0.628 & 0.620 & \cellcolor{c2}0.638 & \cellcolor{c1}0.642\\
\toprule
  \end{tabular}
}
%%%%%%%%%%%
%%%%%%%%%%%
\subfloat[C7-D4]{
\begin{tabular}{ccccc}
\toprule
%$m=2$ & $m=3$ & $m=5$\\
$2$ & $3$ & $5$\\
\midrule
%% Na & Na & Na\\
%% 0.648 & Na & Na\\
0.616 & \cellcolor{c1}0.623 & Na\\
\cellcolor{c1}0.628 & \cellcolor{c2}0.623 & 0.621\\
\cellcolor{c1}0.623 & 0.618 & \cellcolor{c2}0.619\\
0.632 & \cellcolor{c2}0.633 & \cellcolor{c1}0.635\\
 0.618 & \cellcolor{c1}0.630 & \cellcolor{c2}0.627\\
0.593 & \cellcolor{c2}0.596 & \cellcolor{c1}0.604\\
0.600 & \cellcolor{c1}0.605 & \cellcolor{c2}0.602\\
\cellcolor{c1}0.614 & \cellcolor{c2}0.613 & 0.611\\
\toprule
\end{tabular}
}
%%%%%%%%%%%
}
\end{table}

\subsection{Similarity of feature values}
\label{sec:comparison_redu_features}

%We analyze the similarity between features and their dimensionality reduction versions.
%need to understand

\begin{figure}[t]
   \centering
\subfloat[\texttt{ela\_meta.lin\_simple.coef.max}]{
   \includegraphics[width=0.23\textwidth]{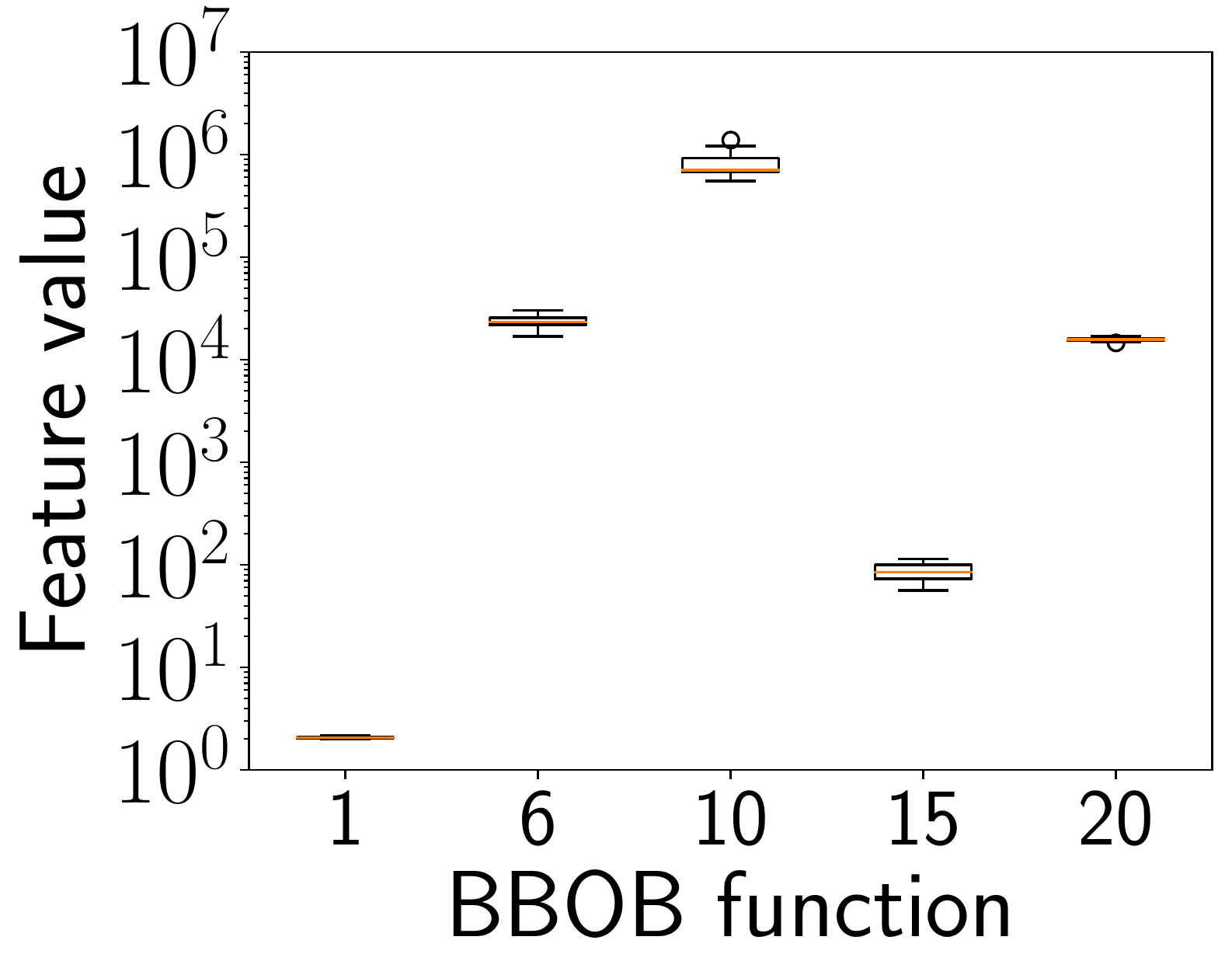}
}
\subfloat[\texttt{d\_ela\_meta.lin\_simple.coef.max}]{
   \includegraphics[width=0.23\textwidth]{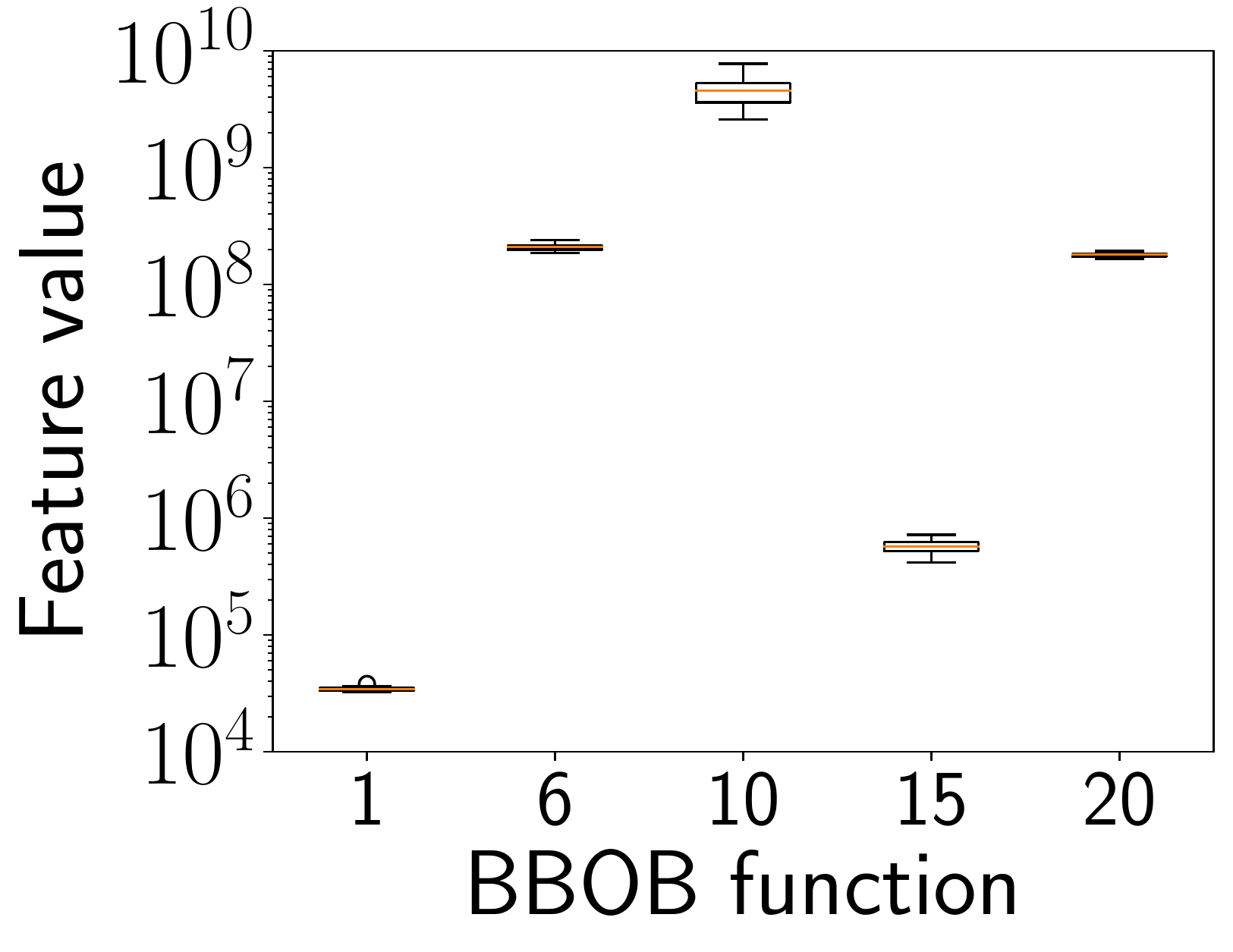}
}
\caption{Boxplots of values of \texttt{ela\_meta.lin\_simple.coef.} \texttt{max} and its dimensionality reduction version on 15 instances of $f_1, f_6, f_{10}, f_{15}$, and $f_{20}$ with $n=160$.}
%\caption{Boxplots of values of \texttt{ela\_meta.lin\_simple.coef.max} and its dimensionality reduction version on $f_1, f_6, f_{10}, f_{15}$, and $f_{20}$ with $n=160$.}
   \label{fig:feature_similarity}
\end{figure}

We analyze the similarity between features and their dimensionality reduction versions.
Figure \ref{fig:feature_similarity} shows values of the \texttt{ela\_meta.lin\_} \texttt{simple.coef.max} feature and its dimensionality reduction version ($m=2$) on 15 instances of $f_1, f_6, f_{10}, f_{15}$, and $f_{20}$ with $n=160$.
We selected them from the five function groups, respectively.
We also selected \texttt{ela\_meta.lin\_simple.coef.max} based on the results in Section \ref{sec:feature_importance}.
Figures S.5--S.130 show values of the other features in C7-D2 and C7-D4.
As seen from the scale of the y-axis in Figures \ref{fig:feature_similarity}(a) and (b), values of the original feature and its dimensionality reduction version are totally different.
%
%This is because PCA cannot reduce the dimensionality without any information loss.
%This is because PCA cannot reduce the dimensionality without any information loss.
The cause of this observation may be the PCA transformation.
An analysis of the proposed framework with nonlinear transformation techniques (e.g., t-SNE \cite{vanderMaatenH08}) is an avenue for future work.
%
%Thus, we \textit{cannot} say that a feature computed by the proposed framework always approximates its original version.
%Thus, we \textit{cannot} say that a feature computed by the proposed framework always approximates its original version.
%While the \texttt{ela\_meta.lin\_simple.coef.max} feature value is in the range $[1, 10^7]$, its dimensionality reduction version is in the range$[10^4, 10^{10}]$.
However, the relative rankings of the feature values on $f_1$, $f_6$, $f_{10}$, $f_{15}$, and $f_{20}$ are similar in Figures \ref{fig:feature_similarity}(a) and (b).
For example, the two features take the smallest and largest values on $f_1$ and $f_{10}$, respectively.

\begin{figure}[t]
   \centering
   \includegraphics[width=0.4\textwidth]{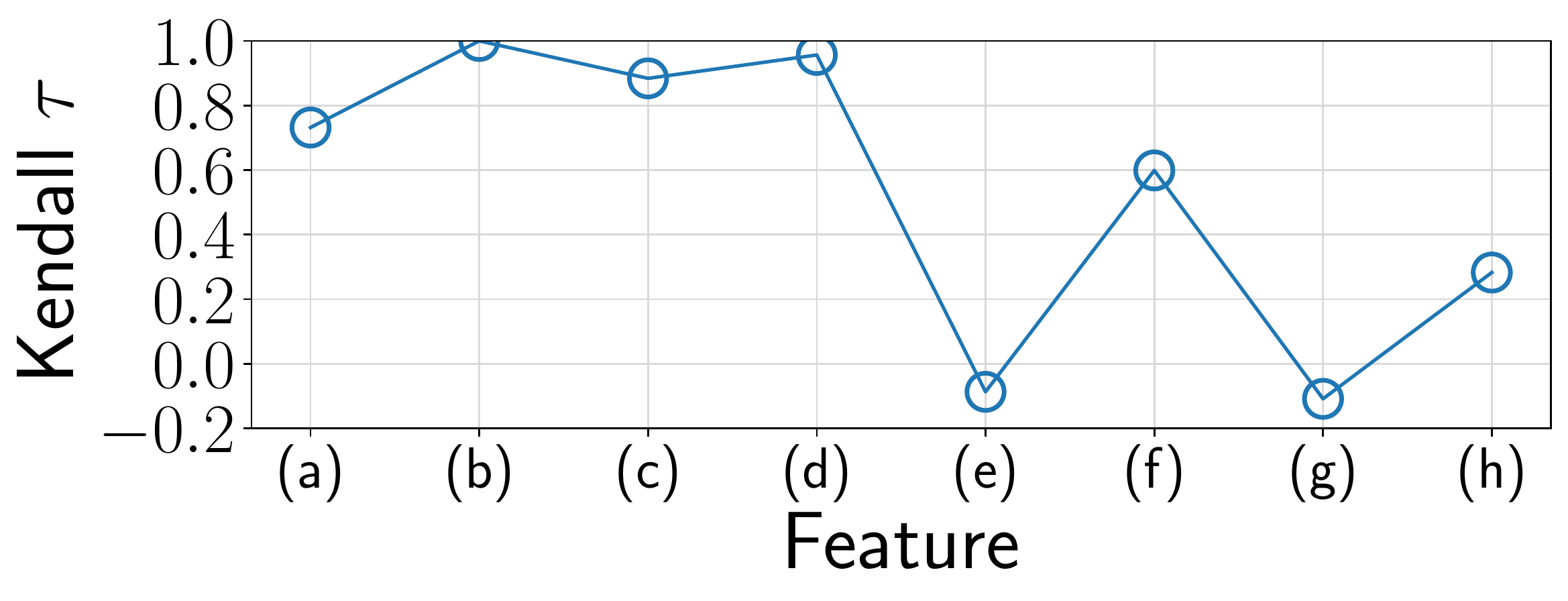}
\caption{Kendall $\tau$ values of \texttt{ela\_meta} and \texttt{d\_ela\_meta} for  $n=160$, where (a) \texttt{lin\_simple.adj\_r2}, (b) \texttt{lin\_simple.intercept}, (c) \texttt{lin\_simple.coef.min}, (d) \texttt{lin\_simple.coef.max}, (e) \texttt{lin\_simple.coef.max\_by\_min}, (f) \texttt{quad\_simple.adj\_r2}, (g) \texttt{quad\_simple.cond}, and (h) \texttt{costs\_runtime}.}
   \label{fig:feature_similarity_kendall}
\end{figure}

%For further investigation, 
Figure \ref{fig:feature_similarity_kendall} shows the Kendall rank correlation coefficient $\tau$ values of the 8 \texttt{ela\_meta} and \texttt{d\_ela\_meta} features on the 24 BBOB functions with $n=160$.
%For other features,
Since 3 out of the 11 \texttt{ela\_meta} and \texttt{d\_ela\_meta} features include non-unique values, we could not calculate their $\tau$ values.
Let $\vector{a} = (a_1, ..., a_{24})^{\top}$ and $\vector{b} = (b_1, ..., b_{24})^{\top}$ be the values of the original feature and its dimensionality reduction version on the 24 BBOB functions respectively, where $a_i \in \vector{a}$ and $b_i \in \vector{b}$ are the average over 15 instances of $f_i$ ($i \in \{1, ..., 24\}$).
The $\tau$ value quantifies the similarity of the rankings of $\vector{a}$ and $\vector{b}$.
Here, $\tau$ takes a value in the range $[-1,1]$.
If the rankings of $\vector{a}$ and $\vector{b}$ are perfectly consistent, $\tau=1$.
If they are perfectly inconsistent, $\tau=-1$.
If they are perfectly independent, $\tau=0$.

%% If the rankings of $\vector{a}$ and $\vector{b}$ are perfectly inconsistent, $\tau=-1$.
%% If the rankings of $\vector{a}$ and $\vector{b}$ are perfectly independent, $\tau=0$.

%Since $\tau = 0.96$ as shown in Figure \ref{fig:feature_similarity_kendall}, \texttt{ela\_meta.lin\_simple.coef.max} is strongly consistent with \texttt{d\_ela\_meta.lin\_simple.coef.max} ($\tau = 0.96$).

As shown in Figure \ref{fig:feature_similarity_kendall}, \texttt{ela\_meta.lin\_simple.coef.max} is strongly consistent with \texttt{d\_ela\_meta.lin\_simple.coef.max} ($\tau = 0.96$).
We can see high $\tau$ values for the following three features: \texttt{lin\_} \texttt{simple.adj\_r2}, \texttt{lin\_simple.intercept}, and \texttt{lin\_simple.coef.} \texttt{min}.
These results indicate that the relative rankings of values of some features and their dimensionality reduction versions can possibly be similar even if their absolute values are different.
We believe that a feature computed by the proposed framework can substitute for its original version when their relative rankings are similar.

%We believe that features computed by the proposed framework were helpful for high-level property classification in Section \ref{sec:comparison_dimredu_e2} for this reason.

Note that this is not always the case.
Since $\tau$ is close to zero for \texttt{lin\_simple.coef.max\_by\_min}, \texttt{quad\_simple.cond}, and \texttt{costs\_} \texttt{runtime} (see (e), (g), and(h) in Figure \ref{fig:feature_similarity_kendall}), their original and dimensionality reduction versions are almost independent.
Figures S.131--S.135 show results of the other features.
The results of \texttt{d\_ela\_level} and \texttt{d\_gcm} are relatively similar to Figure \ref{fig:feature_similarity_kendall}.
In contrast, we observed that $\tau$ is close to zero for most features in the \texttt{d\_cm\_angle}, \texttt{d\_cm\_grad}, and \texttt{d\_cm\_conv} classes.
It would be better to consider that the dimensionality reduction versions of these features are not related to their original versions.

%Thus, we should consider a dimensionality reduction version of such a featrure is completely different from its original version.

% the Kendall rank correlation coefficient $\tau$

%% file: conclusion.tex
\section{Conclusion}
\label{sec:conclusion}

% (see \ref{sec:cputime})

%First, we analyzed the computation time of the ELA features for large dimensions.

First, we pointed out the computational cost issue in the ELA features for large dimensions.
%Since the beginning of this research field, \texttt{ela\_} \texttt{level} and \texttt{ela\_meta} have been recognized as important feature classes.
Our results revealed that the two important feature classes (\texttt{ela\_level} and \texttt{ela\_meta}) cannot be applied to large-scale optimization due to their high computational cost.
%As pointed out in \cite{KerschkeT2019flacco}, the cell mapping feature classes are also computable only for small dimensions.
%
To address this issue, we proposed the dimensionality reduction framework that computes features in a reduced low-dimensional space.
Our results show that the proposed framework can drastically reduce the computation time of \texttt{ela\_level} and \texttt{ela\_meta}.
Our results also show the effectiveness of features (including the four cell mapping features) computed by the proposed framework on the BBOB functions with up to 640 dimensions.
In addition, we found that the relative rankings of values of some (not all) features and their dimensionality reduction versions are similar.

%Our results also show that some features computed by the proposed framework can substitute for their original versions on the BBOB functions with up to $n=640$.

%We believe that this work is the first step toward ELA for large-scale optimization.

We focused on extending the existing features for large dimensions, but it is promising to design a new computationally cheap feature.
An extension of landscape-aware algorithm selection methods for large-scale optimization is an avenue for future work.
It is not obvious if the classification accuracy shown in this paper is good enough for practical purposes.
There is room for discussion about the benchmarking methodology of the ELA approach.